\documentclass[letterpaper,journal]{IEEEtran}

\usepackage{amsmath,amsfonts,amssymb}
\usepackage{array}
\usepackage[caption=false,font=normalsize,labelfont=sf,textfont=sf]{subfig}
\usepackage{textcomp}
\usepackage{stfloats}
\usepackage{placeins}
\usepackage{url}
\usepackage{graphicx}
\usepackage{cite}
\usepackage{booktabs}
\usepackage{multirow}
\usepackage{makecell}
\usepackage[table]{xcolor}
\usepackage{tikz}
\usetikzlibrary{arrows.meta,positioning,calc,fit,backgrounds}
\usepackage{algorithm}
\usepackage{algorithmic}
\usepackage{etoolbox}
\BeforeBeginEnvironment{equation}{\ifhmode\par\fi}
\AtBeginEnvironment{equation}{%
  \fontsize{9pt}{10pt}\selectfont
  \setlength{\abovedisplayskip}{4pt plus 1pt minus 1pt}%
  \setlength{\belowdisplayskip}{4pt plus 1pt minus 1pt}%
  \setlength{\abovedisplayshortskip}{2pt plus 1pt minus 1pt}%
  \setlength{\belowdisplayshortskip}{4pt plus 1pt minus 1pt}%
}

\newcommand{\ours}{TC-ADA}

\newcommand{\source}{\mathcal{S}^{l}}
\newcommand{\targetl}{\mathcal{T}^{l}}
\newcommand{\targetu}{\mathcal{T}^{u}}
\newcommand{\target}{\mathcal{T}}

\newcommand{\cmark}{$\checkmark$}
\definecolor{rankone}{RGB}{190,224,241}
\definecolor{ranktwo}{RGB}{248,215,158}
\definecolor{rankthree}{RGB}{226,226,226}
\newcommand{\best}[1]{\cellcolor{rankone}\textbf{#1}}
\newcommand{\second}[1]{\cellcolor{ranktwo}#1}
\newcommand{\third}[1]{\cellcolor{rankthree}#1}
\newcommand{\tablenotesep}{\par\vspace{0.8mm}}

\begin{document}

\title{TC-ADA: One-Shot Active Domain Adaptation for Semantic Segmentation}

\author{Weihao Yan,
Yeqiang Qian,~\IEEEmembership{Member,~IEEE,}
    Yueyuan Li, Tao Li, \\
Chunxiang Wang,~\IEEEmembership{Member,~IEEE,}
and Ming Yang,~\IEEEmembership{Member,~IEEE}
\thanks{This work is supported by the National Natural Science Foundation of China under Grant 62473253. \emph{(Corresponding authors: Yeqiang Qian; Ming Yang.)}}
\thanks{Weihao Yan, Yeqiang Qian, Yueyuan Li, Tao Li, Chunxiang Wang and Ming Yang are with the School of Automation and Intelligent Sensing, Shanghai Jiao Tong University, Key Laboratory of System Control and Information Processing, Ministry of Education of China, Shanghai, 200240, China (email: qianyeqiang@sjtu.edu.cn; mingyang@sjtu.edu.cn).}%
}

\markboth{IEEE Transactions Manuscript, 2026}{Yan \MakeLowercase{\textit{et al.}}: TC-ADA: One-Shot Active Domain Adaptation}

\maketitle

\begin{abstract}
Manual dense annotation remains a major obstacle to deploying semantic segmentation models in new driving environments. Active domain adaptation (ADA) seeks label-efficient transfer by annotating only a selected portion of the target domain. Existing ADA methods commonly implement this process through multiple rounds of acquisition, annotation, and retraining. We study a practical one-shot image-level setting that selects and densely annotates a fixed target subset in a single round, followed by uninterrupted adaptation. Within this setting, we develop Target-Calibrated Active Domain Adaptation (TC-ADA) as a joint design of complete-image acquisition and target-calibrated adaptation. Stage~1 uses visual representations from a vision foundation model (VFM) together with semantic predictions from a fixed unsupervised domain adaptation model to select representative and informative target images without target annotations. Stage~2 jointly uses labeled source data, labeled target data, and the remaining unlabeled target data, while calibrating source and target supervision under limited target labels. Extensive experiments across five synthetic-to-real and real-to-real driving transfers show consistent improvements over representative ADA baselines. With only 23 to 46 labeled target images on four transfers and 140 on Mapillary, TC-ADA stays within 1.9 mean intersection over union (mIoU) points of target-only full supervision. Code will be available at \url{https://github.com/ywher/TC-ADA}.
\end{abstract}


\section{Introduction}
\IEEEPARstart{I}{mage} semantic segmentation is a fundamental perception task that enables intelligent vehicles and robots to understand their surroundings~\cite{adaptiveseg,open_tcsvt,bssnet,adaptiveocc}. Current semantic segmentation methods typically rely on fully supervised learning with large-scale dense manual annotations. Acquiring such annotations for every deployment domain is costly and impractical.

\begin{figure}[!t]
    \centering
    \includegraphics[width=1.0\linewidth]{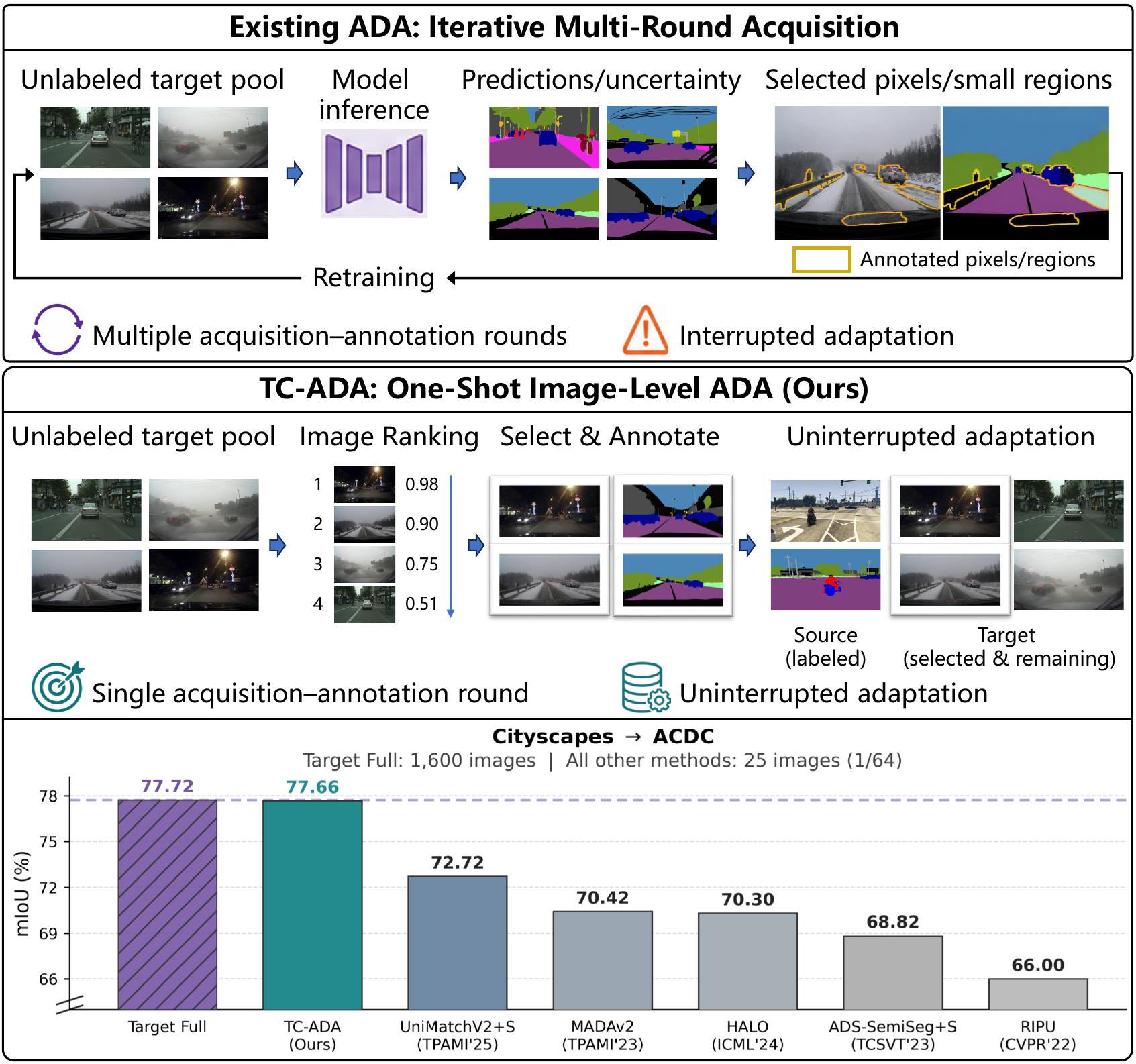}
    \caption{Motivation and protocol overview of \ours{}. It replaces iterative acquisition--annotation--retraining cycles with a single complete-image annotation round followed by uninterrupted adaptation. \ours{} reaches 77.66\% mean intersection over union (mIoU) using only 25 of 1,600 ACDC target images, close to the 77.72\% full-supervision reference.}
    \label{fig:intro_teaser}
\end{figure}

Active domain adaptation (ADA) offers a promising way to alleviate the annotation burden by selectively labeling informative target data~\cite{tcsvt_active,ripu,d2ada,halo,ning2023madav2,dwba,bada}. However, most existing ADA methods for semantic segmentation rely on iterative multi-round acquisition. They repeatedly perform target-pool inference, human annotation, and model retraining as adaptation proceeds. Although earlier annotations can benefit later acquisition, these repeated cycles increase operational overhead and interrupt the training workflow. 

We therefore study \emph{one-shot image-level ADA}, where all target annotations are collected once before adaptation begins. A fixed subset of complete target images is selected and densely annotated in this single round, after which the model is adapted without requesting additional annotations. This protocol avoids repeated inference--annotation--retraining cycles. However, building an effective one-shot ADA system involves two key challenges. Before target labels are available, the main challenge is to select a representative and informative target subset. After annotation, the challenge is to effectively use labeled source data, labeled target data, and the remaining unlabeled target data for adaptation.

To address these challenges, we propose Target-Calibrated Active Domain Adaptation (\ours{}). Stage~1 performs one-shot target selection using visual representations from a vision foundation model (VFM) and semantic predictions from a fixed unsupervised domain adaptation (UDA) model. It aims to select target images that are both representative and informative without using target annotations. Stage~2 jointly uses labeled source data, labeled target data, and the remaining unlabeled target data for adaptation. It adjusts the contributions of source and target supervision under limited target labels, while a target-aware initialization provides a better starting point for adaptation. Fig.~\ref{fig:intro_teaser} summarizes the difference between iterative ADA and our TC-ADA.

Extensive experiments are conducted on five driving adaptation settings covering synthetic-to-real, adverse-weather, and geographic domain shifts. We evaluate segmentation accuracy, acquisition efficiency, and model generalization against representative ADA methods for semantic segmentation. The main contributions are summarized as follows:
\begin{itemize}
\item We develop TC-ADA, a practical one-shot image-level ADA framework for semantic segmentation. It performs complete-image annotation in a single round, followed by uninterrupted adaptation without further annotations.

\item We propose a target selection strategy for representative and informative images without target labels. It combines VFM-based coverage with UDA-derived semantics to reduce redundancy and better capture important semantics.

\item We design target-calibrated adaptation using labeled source, labeled target, and unlabeled target data. It balances source and target supervision under limited target labels for more effective adaptation.

\item Experiments across five driving benchmarks show consistent improvements over representative ADA baselines under controlled image-level settings. At the lowest label budgets, \ours{} remains within 1.9 mIoU points of target-only full supervision on all five transfers, with an average gap of 1.07 points. Code will be available at \url{https://github.com/ywher/TC-ADA}.
\end{itemize}

\section{Related Work}

\subsection{Unsupervised Domain-Adaptive Semantic Segmentation}
Unsupervised domain adaptation (UDA) transfers knowledge from labeled source to unlabeled target data through feature/output alignment, self-training, or cross-domain augmentation. DACS~\cite{dacs} established cross-domain ClassMix as a strong self-training paradigm. DAFormer~\cite{daformer} combined transformer representations with a robust adaptation recipe, HRDA~\cite{hrda} reconciled global context and high-resolution detail, and MIC~\cite{mic} enforced consistency under masked target views. Reliability-aware and context-preserving objectives further improve class-wise pseudo supervision and cross-domain composition~\cite{tufl,CAM}. Segmentation foundation models can also refine target pseudo labels~\cite{sam4udass}, while SCSD~\cite{scsd} further studies the one-shot UDA setting.
Recent UDA methods build adaptation systems around self-supervised VFMs. DINOv2 and DINOv3 provide transferable patch representations~\cite{dinov2,dinov3}. Rein and Rein++ adapt frozen encoders through lightweight modules~\cite{rein,reinpp}, and VFM4UDASS revisits UDA architecture and optimization for these features~\cite{vfmuda}. 

Despite these advances, UDA relies on potentially noisy target predictions without target labels, especially under large domain gaps. TC-ADA further selects a small set of complete target images and adapts with labeled source, selected target, and remaining unlabeled target data. 

\subsection{Semi-Supervised Semantic Segmentation}
Semi-supervised semantic segmentation learns from a small labeled subset and a larger unlabeled pool from the same domain~\cite{semi_survey}. Its dominant paradigm combines pseudo labeling, consistency regularization, and strong augmentation. ClassMix constructs class-level mixed views~\cite{dacs}, UniMatch strengthens weak-to-strong consistency with complementary perturbations~\cite{unimatch}, and UniMatchV2 extends this design to VFM encoders through complementary dropout~\cite{unimatchv2}. Information transfer and knowledge distillation provide additional interactions between labeled and unlabeled samples~\cite{infor_tcsvt,tcsvt_selfsemi}. PerturbMatch~\cite{perturbmatch} dynamically adjusts the perturbation strength over the course of training, improving the stability and effectiveness of unlabeled-data utilization.
Additionally, AllSpark transfers complementary unlabeled features into the labeled stream~\cite{allspark}, while SemiVL introduces vision-language priors~\cite{semivl}. Our prior work SSMIS shows that distribution-aware density-k-center (DKC) sampling can improve labeled-subset coverage beyond random selection in semi-supervised medical segmentation~\cite{ssmis}.

ADS-SemiSeg~\cite{adssemiseg} learns from a predefined labeled subset through adversarial dual students, while DIP~\cite{dip} learns prototypes from given target support. TC-ADA instead selects that support from the unlabeled pool and then adapts with source, selected-target, and remaining unlabeled-target data.

\subsection{Active Domain Adaptation for Semantic Segmentation}
Active learning reduces annotation by selecting informative images, regions, or pixels. Representative strategies cover the feature space with k-center~\cite{k-center}, combine gradient uncertainty and diversity in BADGE~\cite{badge}, or favor representative clusters at low budgets in TypiClust~\cite{typiclust}. DCoV instead uses confidence variation over training to construct generalized coresets~\cite{dcov}. These methods motivate target selection, while ADA must additionally account for cross-domain shift.

RIPU combines regional impurity and uncertainty~\cite{ripu}, D2ADA dynamically balances uncertainty and target density~\cite{d2ada}, and HALO performs pixel acquisition in a hyperbolic representation~\cite{halo}. More recent methods further improve fine-grained acquisition: DWBA-ADA emphasizes minority classes and category boundaries through dynamic weighting and boundary-aware selection~\cite{dwba}, while BADA selects samples around decision and domain-shift boundaries under adverse conditions~\cite{bada}. These iterative methods use updated models to guide subsequent queries, but require repeated acquisition, annotation, and retraining cycles. Gao et al.~\cite{gao_superpixel} further reduce fine-grained annotation cost through uncertainty-guided superpixel fusion and information-rich superpixel selection. However, the method remains sensitive to superpixel quality, which can introduce semantic inconsistency and boundary errors.
MADAv2 is an important single-round image-level alternative that uses multiple anchors to represent multimodal source and target distributions~\cite{ning2023madav2}.

One-shot acquisition faces the challenge of selecting a representative and informative target subset without target labels. This is particularly difficult under domain shift, where source-trained uncertainty may be unreliable and complete-image selection can suffer from substantial semantic redundancy. Semi-supervised domain adaptation (SSDA), by contrast, assumes that a labeled target subset is already given and improves only the adaptation stage through consistency, mixing, self-training, or vision-language guidance~\cite{semidavil}. \ours{} addresses both parts: it selects complete target images once, then uses an SSDA-like data configuration for uninterrupted adaptation. 

\section{Method}

\subsection{Problem Definition}
Let $\source=\{(x_i^s,y_i^s)\}_{i=1}^{N_s}$ be a labeled source set and $\target=\{x_i^t\}_{i=1}^{N_t}$ an unlabeled target training pool. Under an image-level annotation budget $K\ll N_t$, the acquisition function selects $\mathcal{A}\subset\{1,\ldots,N_t\}$ with $|\mathcal{A}|=K$. 
Dense annotation then produces $\targetl=\{(x_i^t,y_i^t)\}_{i\in\mathcal{A}}$, while the unselected images form $\targetu=\{x_i^t\}_{i\notin\mathcal{A}}$. The final segmentor $f_\theta$ is evaluated on a disjoint target validation set.

We consider one-shot image-level ADA: $\mathcal{A}$ is determined before Stage~2 adaptation, all $K$ selected target images are densely annotated in a single round, and no additional target annotation is performed afterward. The data configuration after annotation resembles SSDA, but the labeled target subset is selected by the method rather than provided as a fixed or random subset. This distinction makes acquisition quality directly affect the final adaptation performance.

\begin{figure*}[!t]
    \centering
    \includegraphics[width=1.0\linewidth]{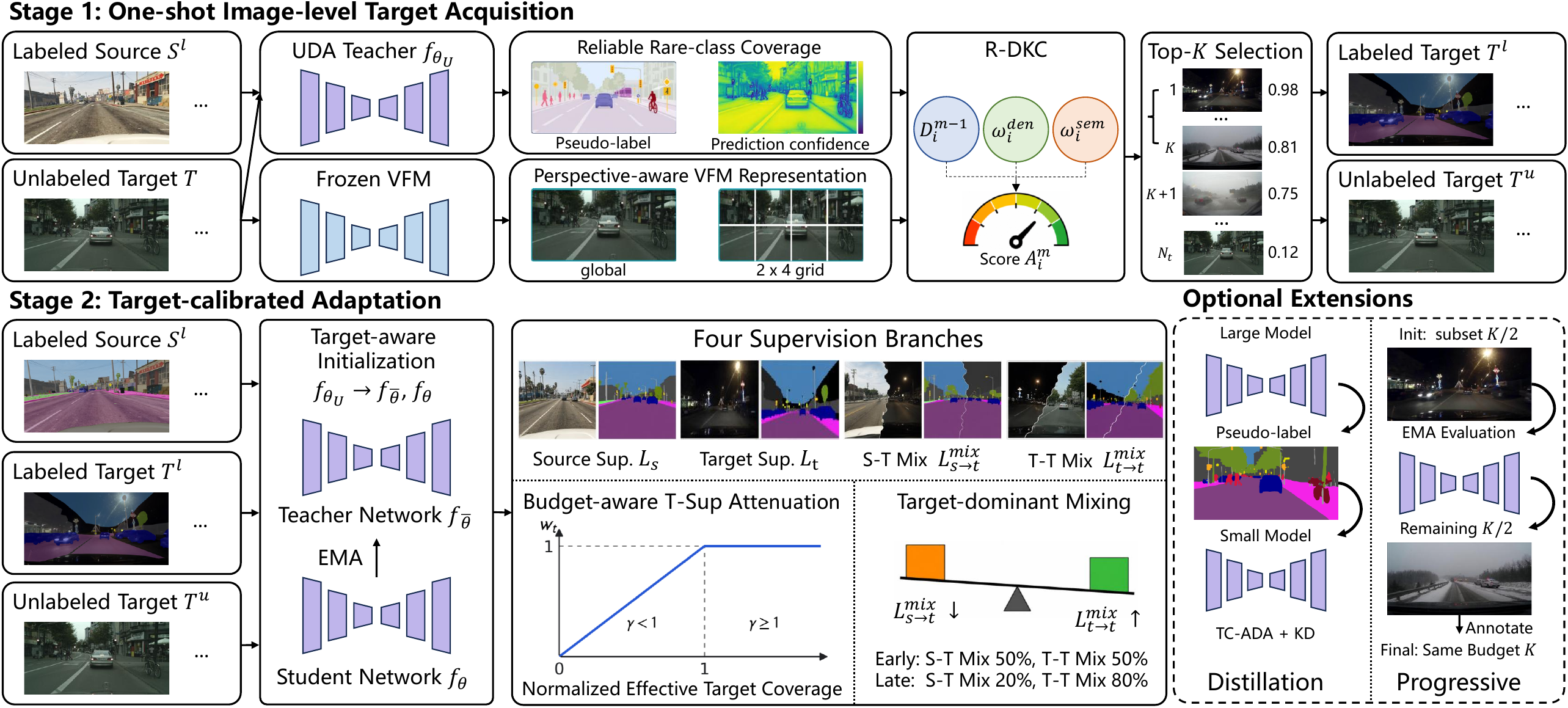}
    \caption{Overview of \ours{}. Stage~1 ranks complete target images using
    perspective-aware VFM features, density--diversity coverage, and reliable
    rare-class evidence. Stage~2 combines source and target supervision through
    two mixed branches, with budget-aware target-supervision attenuation,
    target-dominant mixing, and target-aware initialization. Distillation and
    progressive acquisition are optional extensions.}
    \label{fig:framework}
    \vspace{-1em}
\end{figure*}

\subsection{Framework Overview}
Fig.~\ref{fig:framework} summarizes \ours{}. Stage~1 models the feature-space structure of the target pool with a frozen VFM and computes semantic statistics with a fixed UDA teacher $f_{\theta_U}$. Their combination produces a nested image ranking, from which the first $K$ images are annotated. Stage~2 trains on $\source$, $\targetl$, and $\targetu$ through four complementary terms:
source supervision, direct target supervision, hybrid source--target mixing, and target--target mixing. Target-dominant mixing (TDM) gradually shifts the mixed-loss balance toward the target domain, while a UDA checkpoint provides a target-aware initialization. The UDA teacher and the Stage~2 EMA teacher have distinct roles: the former is fixed before acquisition, whereas the latter evolves with $f_\theta$ during adaptation.
This separation defines our default one-shot protocol. Acquisition observes the target pool only through cached VFM features and fixed-teacher predictions, so changing the Stage~2 model or repeating a training run does not alter the annotation request. In this setting, Stage~2 treats the selected labels as ordinary training supervision and requires no further acquisition or annotation; the optional progressive variant is evaluated separately.

\subsection{Teacher and Model Preparation}
Stage~1 requires a frozen VFM that exposes multi-layer patch features. Our default instantiation uses DINOv3~\cite{dinov3}. The Stage~2 segmentor combines the frozen encoder with Rein adapters~\cite{rein} and an HRDA decoder~\cite{hrda}, so only the adapters and decoder are optimized.

The UDA model is trained before acquisition using only source labels and unlabeled target images, following~\cite{hrda}. It supplies target predictions for Stage~1 and target-aware adapter/decoder parameters for Stage~2. Let $C$ be the number of classes and $\Omega_i$ the pixel lattice of target image $x_i^t$. At pixel $u\in\Omega_i$, the fixed UDA teacher produces
\begin{equation}
    p_i^c(u)=\left[\operatorname{softmax}
    \bigl(f_{\theta_U}(x_i^t)(u)\bigr)\right]_c,\quad
    \hat y_i(u)=\operatorname*{argmax}_{c\in\{1,\ldots,C\}}p_i^c(u),
    \label{eq:teacher_prediction}
\end{equation}
where $p_i^c(u)$ and $\hat y_i(u)$ are the class probability and pseudo label.
We also store the pixel confidence
\begin{equation}
    q_i(u)=\max_c p_i^c(u).
    \label{eq:teacher_confidence}
\end{equation}
These predictions are computed once over the target pool and reused for acquisition. We use $\epsilon>0$ as a numerical stabilizer in subsequent normalizations.

\subsection{Stage~1: One-shot Image-level Target Acquisition}

\subsubsection{Perspective-aware VFM representation}
Driving images exhibit a stable perspective layout: distant structures frequently occupy the upper field, road regions dominate the lower field, and small traffic participants appear locally. A global patch average can therefore suppress sparse foreground evidence. We retain global context while preserving local layout through a component-normalized global--grid representation.

Let $F_i^\ell\in\mathbb{R}^{H_p\times W_p\times D_f}$ be the VFM patch feature map from layer $\ell\in\mathcal{L}$, where $H_p$, $W_p$, and $D_f$ denote the patch-grid height, patch-grid width, and feature dimension, respectively.
The region set $\mathcal{R}$ contains the complete patch lattice and the eight cells of a $2\times4$ grid. The grid matches the approximate $1{:}2$ aspect ratio of driving images and yields near-square local cells without the dimensional cost of a very fine pyramid. For region $r\in\mathcal{R}$, we compute
\begin{equation}
    g_i^{\ell,r}=\frac{1}{|r|}\sum_{v\in r}F_i^\ell(v),\qquad
    z_i^{\ell,r}=\frac{g_i^{\ell,r}}
    {\|g_i^{\ell,r}\|_2+\epsilon},
    \label{eq:regional_embedding}
\end{equation}
where $v$ indexes patch tokens. The image representation is
\begin{equation}
    \tilde z_i=\underset{\ell\in\mathcal{L},\,r\in\mathcal{R}}
    {\operatorname{concat}}\bigl(z_i^{\ell,r}\bigr),\qquad
    z_i=\frac{\tilde z_i}{\|\tilde z_i\|_2+\epsilon}.
    \label{eq:image_embedding}
\end{equation}
Normalizing each layer--region component before concatenation prevents a single scale or spatial region from dominating the distance metric.

\subsubsection{Density--diversity coverage}
We combine k-center diversity~\cite{k-center} with local target density. Since $z_i$ is unit-normalized, the image distance can be calculated as $d(i,j)=1-z_i^\top z_j$. The first center is the sample farthest from the normalized target-pool mean. At each subsequent greedy step $m\geq2$, the diversity of an unselected candidate is
\begin{equation}
    D_i^{(m-1)}=\min_{j\in\mathcal{A}_{m-1}}d(i,j),
    \qquad i\notin\mathcal{A}_{m-1},
    \label{eq:kcenter_distance}
\end{equation}
where $\mathcal{A}_{m-1}$ contains the previously selected images. Let $\mathcal{N}_k(i)$ be the $k$ nearest neighbors of $i$ in the embedding space. We estimate local density and convert it to a bounded weight:
\begin{equation}
\begin{aligned}
    \delta_i&=\left(
    \frac{1}{k}\sum_{j\in\mathcal{N}_k(i)}\|z_i-z_j\|_2+\epsilon
    \right)^{-1},\\
    \omega_i^{\mathrm{den}}&=\operatorname{clip}\left(
    \frac{\delta_i}
    {\delta_i+\operatorname{median}_{n}(\delta_n)+\epsilon},
    \omega_{\min}^{\mathrm{den}},\omega_{\max}^{\mathrm{den}}
    \right).
\end{aligned}
\label{eq:density_weight}
\end{equation}
The product $D_i^{(m-1)}\omega_i^{\mathrm{den}}$ favors candidates that add coverage without overselecting isolated outliers.

\subsubsection{Reliable rare-class coverage}
Feature-space coverage alone cannot distinguish images that contain reliably predicted rare classes from images dominated by frequent background classes.
We therefore complement it with one semantic cue: reliable
coverage of underrepresented target classes.

For class $c$, define its predicted area ratio $a_i^c$, the mean confidence $\bar q_i^c$ over pixels assigned to $c$, and its pool-level predicted pixel frequency $f_c$:
\begin{equation}
\begin{aligned}
    a_i^c&=\frac{1}{|\Omega_i|}\sum_{u\in\Omega_i}
    \mathbb{I}[\hat y_i(u)=c],\\
    \bar q_i^c&=
    \frac{\sum_{u\in\Omega_i}\mathbb{I}[\hat y_i(u)=c]q_i(u)}
    {\sum_{u\in\Omega_i}\mathbb{I}[\hat y_i(u)=c]+\epsilon},\\
    f_c&=\frac{\sum_{i=1}^{N_t}\sum_{u\in\Omega_i}
    \mathbb{I}[\hat y_i(u)=c]}
    {\sum_{i=1}^{N_t}|\Omega_i|},
\end{aligned}
\label{eq:class_statistics}
\end{equation}
where $\Omega_i$ is the set of pixel locations in $x_i^t$, $|\Omega_i|$ is its number of pixels, and $\mathbb{I}[\cdot]$ is the indicator function. The normalized rare-class weight and reliable rare-class coverage are
\begin{equation}
    w_c^{\mathrm{rare}}=
    \frac{(f_c+\epsilon)^{-1/2}}
    {C^{-1}\sum_{c'=1}^{C}(f_{c'}+\epsilon)^{-1/2}},
    \label{eq:rare_weight}
\end{equation}

\begin{equation}
    R_i=\sum_{c=1}^{C}\sqrt{a_i^c}\,
    \mathbb{I}[a_i^c\geq\tau_a]\,
    \bar q_i^c w_c^{\mathrm{rare}}.
    \label{eq:rare_utility}
\end{equation}
The area threshold $\tau_a$ suppresses tiny pseudo-label fragments, while the square root limits domination by large regions.

Thus, $R_i$ is large only when an image contains a nontrivial area of a pool-level rare class and the teacher predicts that region reliably. It does not favor uncertainty by itself, which avoids allocating the limited budget to images dominated by noisy pseudo-label boundaries.

For an image-level statistic $\xi_i$, define min--max normalization over the target pool as
\begin{equation}
    \mathcal{N}(\xi_i)=
    \frac{\xi_i-\min_{1\leq n\leq N_t}\xi_n}
    {\max_{1\leq n\leq N_t}\xi_n-
     \min_{1\leq n\leq N_t}\xi_n+\epsilon}.
    \label{eq:pool_normalization}
\end{equation}
The semantic weight is
\begin{equation}
\begin{aligned}
    \tilde\omega_i^{\mathrm{sem}}
    &=1+\lambda_R\mathcal{N}(R_i),\\
    \omega_i^{\mathrm{sem}}
    &=\operatorname{clip}\left(
    \frac{\tilde\omega_i^{\mathrm{sem}}}
    {N_t^{-1}\sum_{n=1}^{N_t}\tilde\omega_n^{\mathrm{sem}}},
    \omega_{\min}^{\mathrm{sem}},\omega_{\max}^{\mathrm{sem}}
    \right),
\end{aligned}
\label{eq:semantic_weight}
\end{equation}
where $\lambda_R\geq0$ controls the strength of rare-class guidance and $\omega_{\min}^{\mathrm{sem}},\omega_{\max}^{\mathrm{sem}}$ bound its influence. The density bounds $\omega_{\min}^{\mathrm{den}},\omega_{\max}^{\mathrm{den}}$ play the same role for local density.
The final reliability-guided density--k-center (R-DKC) score is
\begin{equation}
    A_i^{(m)}=D_i^{(m-1)}
    \omega_i^{\mathrm{den}}\omega_i^{\mathrm{sem}}.
    \label{eq:stage1_score}
\end{equation}
Pool normalization makes the semantic score comparable across target pools, while mean normalization and clipping keep semantic outliers from overwhelming feature-space coverage. At every step, the candidate with the largest $A_i^{(m)}$ is appended to $\mathcal{A}_{m-1}$ and all nearest-center distances are updated. Running the procedure to the largest budget produces one ordered list; smaller budgets are obtained from its prefixes. Thus, VFM features and teacher statistics are computed once, no segmentor is retrained during selection, and the same ranking applies across budgets and adaptation runs.

\subsection{Stage~2: Target-calibrated Adaptation}

\subsubsection{Pseudo supervision and mixed samples}
Stage~2 maintains an exponential moving average (EMA) teacher $f_{\bar\theta}$ for the unlabeled target set. After iteration $\tau$, its parameters are updated from the student network as
\begin{equation}
    \bar\theta_\tau=
    \alpha_\tau\bar\theta_{\tau-1}+
    (1-\alpha_\tau)\theta_\tau,\qquad
    \alpha_\tau=\min\left(1-\frac{1}{\tau+1},\alpha\right),
    \label{eq:ema_update}
\end{equation}
where $\alpha$ is the maximum EMA momentum. For an unlabeled target image $x_j^u$, let $\bar p_j^{u,c}(u)=[\operatorname{softmax}(f_{\bar\theta}(x_j^u)(u))]_c$ denote the teacher probability of class $c$ at pixel $u$. The pseudo label and confidence are $\bar y_j^u(u)=\operatorname*{argmax}_c\bar p_j^{u,c}(u)$ and $\bar q_j^u(u)=\max_c\bar p_j^{u,c}(u)$, respectively. Following DACS~\cite{dacs}, its reliability is summarized by the fraction of confident pixels,
\begin{equation}
    \eta_j^u=\frac{1}{|\Omega_j|}
    \sum_{u\in\Omega_j}
    \mathbb{I}[\bar q_j^u(u)\geq\tau_p],
    \label{eq:pseudo_reliability}
\end{equation}
where $\tau_p$ is the pseudo-label threshold. The pseudo-labeled target region is weighted by $\eta_j^u$, whereas ground-truth regions have unit weight. This image-adaptive weight reduces the contribution of uncertain target views without discarding their spatial context.

We use the same ClassMix construction for all mixed branches. Let $(x^a,y^a)$ be a labeled donor and $(x^b,y^b)$ a labeled or pseudo-labeled context image. A binary mask $M$ is formed from a random subset of semantic classes present in $y^a$. The mixed sample, label, and pixel weight are
\begin{equation}
\begin{aligned}
    \tilde x&=M\odot x^a+(1-M)\odot x^b,\\
    \tilde y&=M\odot y^a+(1-M)\odot y^b,\\
    \tilde w&=M+(1-M)\eta^b,
\end{aligned}
    \label{eq:classmix_construction}
\end{equation}
where $\odot$ is element-wise multiplication and $\eta^b=1$ for a labeled context or Eq.~\eqref{eq:pseudo_reliability} for an unlabeled one. Standard color and blur perturbations are applied after composition. This common operator makes the difference between branches depend on their supervision sources rather than on unrelated augmentation choices.

\subsubsection{Adaptation objective}
After annotation, Stage~2 samples mini-batches from $\source$, $\targetl$, and $\targetu$. The final objective is
\begin{equation}
    \mathcal{L}_{\mathrm{ADA}}(\tau)=
    \mathcal{L}_s+
    w_t\mathcal{L}_t+
    w_{st}(\tau)\mathcal{L}_{s\rightarrow t}^{\mathrm{mix}}+
    w_{tt}(\tau)\mathcal{L}_{t\rightarrow t}^{\mathrm{mix}},
    \label{eq:ssda_loss}
\end{equation}
where $\tau$ is the training iteration. $\mathcal{L}_s$ and $\mathcal{L}_t$ are supervised segmentation losses on source and selected target images, respectively. The hybrid source--target branch is
\begin{equation}
    \mathcal{L}_{s\rightarrow t}^{\mathrm{mix}}=
    \tfrac{1}{2}\mathcal{L}_{\mathrm{mix}}(\source,\targetl)+
    \tfrac{1}{2}\mathcal{L}_{\mathrm{mix}}(\source,\targetu),
    \label{eq:hybrid_source_mix}
\end{equation}
where $\mathcal{L}_{\mathrm{mix}}$ denotes the pixel-weighted cross-entropy on the mixed sample in Eq.~\eqref{eq:classmix_construction}. ClassMix~\cite{dacs} pastes source classes into labeled or unlabeled target contexts. The two terms use ground truth and EMA pseudo labels on their target portions, respectively. With batch size greater than one, half of the source samples are assigned to each context; with batch size one, the two choices alternate across iterations.

The target--target branch $\mathcal{L}_{t\rightarrow t}^{\mathrm{mix}}$ uses selected labeled target images as donors and unlabeled target images as contexts. It propagates trusted target semantics into strongly augmented views of the unlabeled pool without reintroducing source appearance. Together, the hybrid branch answers what source knowledge remains useful, whereas the target--target branch stabilizes the target decision boundary.

\subsubsection{Budget-aware target-supervision attenuation}
Direct target supervision is accurate, but under an extreme image budget the same small labeled set is revisited throughout a fixed training schedule. Assigning it a constant unit weight can therefore overfit a few scenes and their class distribution. We attenuate only this direct branch according to the target annotation budget. Let $N_l=|\targetl|$ and $N_s=|\source|$. Using the saturated effective count $E_\beta(n)$, we set
\begin{equation}
    E_\beta(n)=\frac{1-\beta^n}{1-\beta}, \quad
    \gamma=\frac{E_\beta(N_l)}{r_0E_\beta(N_s)},\qquad
    w_t=\min(1,\gamma),
    \label{eq:target_sup_attenuation}
\end{equation}
where $\beta\in[0,1)$ controls count saturation, $r_0>0$ sets the reference target-to-source coverage, and $\gamma$ is the normalized effective target coverage. Thus, $w_t<1$ only when the annotated target set is extremely small, and the standard objective is recovered once target coverage is sufficient. The coefficient is resolved once from the annotated target set and introduces no additional model or training state.

\subsubsection{Target-dominant mixing}
Uniformly weighting the two mixed branches can preserve source-dominated gradients late in training. We instead keep their total contribution fixed while progressively favoring target-only mixing. Let $T$ be the total number of Stage~2 training iterations and $\pi_\tau=\min(1,\tau/T)$ the normalized training progress. The target share and the branch weights are
\begin{equation}
    \rho(\tau)=\rho_0+(\rho_1-\rho_0)\pi_\tau,
    \label{eq:target_share}
\end{equation}

\begin{equation}
    w_{tt}(\tau)=B\rho(\tau),\qquad
    w_{st}(\tau)=B\bigl(1-\rho(\tau)\bigr),
    \label{eq:target_dominant_mix}
\end{equation}
where $B$ is the fixed total mixed-loss weight and $0\leq\rho_0\leq\rho_1\leq1$. We set $B=2$, $\rho_0=0.5$, and $\rho_1=0.8$, so training begins with equal mixed-branch weights. Increasing $\rho(\tau)$ transfers weight from source--target to target--target mixing without changing their total contribution or the source-supervised loss scale. TDM calibrates where mixed supervision comes from rather than introducing an additional loss.

\subsubsection{Target-aware initialization}
The fixed UDA model provides a target-aware starting point for Stage~2. For VFM-based segmentors, we transfer its parameter-efficient fine-tuning (PEFT) adapter and decoder while keeping the backbone frozen. 
This is parameter transfer rather than continued UDA training: Stage~2 optimization restarts with a new objective and uses no target annotations for initialization. The checkpoint is the same model used to compute Stage~1 target statistics, so its reuse requires no initialization-only optimization. Warm starts are also common in ADA: D2ADA uses supervised or UDA warm-up checkpoints, while MADAv2 uses a source-warm-up model~\cite{d2ada,ning2023madav2}. Thus, although pre-annotation UDA preparation has a cost, our initialization introduces no separate training run.
Algorithm~\ref{alg:tc_ada} summarizes acquisition and Stage~2 adaptation of one-shot \ours{}.

\begin{algorithm}[t]
\caption{One-shot \ours{}}
\label{alg:tc_ada}
\begin{algorithmic}[1]
\STATE Train the fixed UDA model and cache target predictions.
\STATE Extract component-normalized VFM embeddings for $\target$.
\STATE Greedily rank $\target$ using Eq.~\eqref{eq:stage1_score}; annotate the
first $K$ images to form $\targetl$; let the remaining images form $\targetu$.
\STATE Resolve the direct target-supervision weight with
Eq.~\eqref{eq:target_sup_attenuation}.
\STATE Load the UDA adapter/decoder and initialize the EMA teacher and student network.
\FOR{$\tau=0,\ldots,T-1$}
    \STATE Sample $\source$, $\targetl$, and $\targetu$; generate target pseudo
    labels and confidence weights.
    \STATE Construct hybrid source--target and target--target ClassMix batches.
    \STATE Compute Eq.~\eqref{eq:ssda_loss} using $w_t$ in Eq.~\eqref{eq:target_sup_attenuation} and the TDM weights in
    Eq.~\eqref{eq:target_dominant_mix}.
    \STATE Update the student network and EMA teacher.
\ENDFOR
\end{algorithmic}
\end{algorithm}

\subsubsection{Optional offline output distillation}
Real-time inference is important in driving applications, yet directly adapting a compact segmentor such as PIDNet-S~\cite{pidnet} leaves a sizable accuracy gap from a high-capacity VFM segmentor. We therefore add an optional, deployment-oriented knowledge distillation (KD) step: the strongest DINOv3-L+Rein+HRDA model provides offline pseudo labels for the lightweight student on unlabeled target images. Let $p_i^{\mathrm H,c}(u)$ be the cached probability from this high-capacity teacher and $q_i^{\mathrm H}(u)=\max_c p_i^{\mathrm H,c}(u)$. The student uses
\begin{equation}
    (\tilde y_i(u),\tilde w_i(u))=
    \begin{cases}
        \bigl(\operatorname*{argmax}_c p_i^{\mathrm H,c}(u),
        \omega_{\mathrm H}\bigr),
        & q_i^{\mathrm H}(u)\geq\tau_{\mathrm{KD}},\\
        \bigl(\hat y_i^{\mathrm E}(u),w_i^{\mathrm E}(u)\bigr),
        & \text{otherwise},
    \end{cases}
    \label{eq:offline_kd}
\end{equation}
where $\tau_{\mathrm{KD}}$ is the offline-teacher confidence threshold, $\omega_{\mathrm H}$ is the training weight of an accepted offline pseudo label, and $(\hat y_i^{\mathrm E},w_i^{\mathrm E})$ is the pseudo label and training weight supplied by the online EMA teacher.
The fused labels replace EMA labels only in unlabeled-target regions of the two mixed branches. Teacher inference is performed once, the selected target subset is unchanged, and the teacher is absent during student optimization and deployment. We evaluate this extension separately from the core method.

\subsubsection{Optional Progressive Acquisition}
TC-ADA uses one-shot acquisition by default: all $K$ target images are selected before adaptation and annotated in a single round. To examine the accuracy--cost trade-off of iterative feedback, we additionally consider an optional two-round extension in our experiments. The offline selector first acquires $K/2$ images. After a $T_w$-iteration warm-up, the EMA model re-ranks the remaining target pool and acquires another $K/2$ images. Training then continues without resetting the model or optimizer, while the final annotation budget and total training iterations remain unchanged. This extension requires an additional acquisition--annotation round and is therefore excluded from the default TC-ADA protocol and main comparisons.


\section{Experiments}

\subsection{Datasets and Settings}


We evaluate five driving ADA settings. GTA-to-Cityscapes and SYNTHIA-to-Cityscapes measure synthetic-to-real (S2R) transfer under 19-class and shared 16-class protocols, respectively~\cite{cityscapes,playingfordata,synthia}. Cityscapes-to-ACDC~\cite{acdc}, Cityscapes-to-MUSES~\cite{muses}, and Cityscapes-to-Mapillary~\cite{mapillary} measure real-to-real (R2R) transfer across adverse weather, illumination, and geographic diversity. Table~\ref{tab:datasets} summarizes datasets and split sizes. We abbreviate these transfers as G2C, S2C, C2A, C2Mu, and C2Map, respectively. Segmentation accuracy is reported as mIoU (\%); differences are stated in mIoU points.

\begin{table}[htbp]
\centering
\caption{Driving adaptation datasets and split sizes.}
\label{tab:datasets}
\setlength{\tabcolsep}{3.0pt}
\begin{tabular}{c|l|c|rr|c}
\toprule
Shift & Transfer & Cls. & \#Source & \#Target & \#Val \\
\midrule
\multirow{2}{*}{S2R}
& GTA $\rightarrow$ Cityscapes & 19 & 24,966 & 2,975 & 500 \\
& SYNTHIA $\rightarrow$ Cityscapes & 16 & 9,400 & 2,975 & 500 \\
\midrule
\multirow{3}{*}{R2R}
& Cityscapes $\rightarrow$ ACDC & 19 & 2,975 & 1,600 & 406 \\
& Cityscapes $\rightarrow$ MUSES & 19 & 2,975 & 1,500 & 250 \\
& Cityscapes $\rightarrow$ Mapillary & 19 & 2,975 & 18,000 & 2,000 \\
\bottomrule
\end{tabular}
\vspace{-2em}
\end{table}

\subsection{Implementation Details}
Unless stated otherwise, \ours{} uses a frozen DINOv3-B backbone with Rein adapters and HRDA, optimizing only the adapters and decoder. Stage~1 concatenates component-normalized features from four VFM layers over the whole image and a $2\times4$ grid. It uses $k=20$, $\lambda_R=1$, $\tau_a=0.001$, and density/semantic clipping ranges of $[0.3,1]$/$[0.5,2]$. Stage~2 equally samples labeled and unlabeled target contexts in the hybrid source--target branch and attenuates direct target supervision with $(\beta,r_0)=(0.99,0.5)$. It keeps the total mixed-loss weight at 2 and changes the target--target share from $0.5$ to $0.8$. The UDA checkpoint initializes only the adapters and decoder; all optimization states are reset. 

Models are trained for 40k iterations on $1024\times1024$ crops with a batch size of two and AdamW (adapter learning rate $10^{-4}$, decoder multiplier $10$, and weight decay 0.05). The EMA momentum and pseudo-label threshold are 0.999 and 0.968. Source rare class sampling (RCS) is enabled only for synthetic-to-real transfers~\cite{daformer}. Controlled baselines retain their method-specific acquisition and optimization. Remaining settings follow the released configurations.
Offline distillation uses $(\tau_{\mathrm{KD}},\omega_{\mathrm H})=(0.95,0.95)$.
Where applicable, blue, orange, and gray mark the top three comparable values; maxima are also boldfaced. Reported averages are computed from unrounded scores.
Code will be made publicly available at \url{https://github.com/ywher/TC-ADA}.

\subsection{Comparison with Existing Methods}
\begin{table*}[!t]
\centering
\caption{Protocol-aware comparison at the lowest target-label budgets (mIoU, \%). Parentheses give labeled-image counts for image-level protocols.}
\label{tab:method_compare}
\setlength{\tabcolsep}{2.5pt}
\renewcommand{\arraystretch}{0.98}
\begin{tabular}{l|c|l|cc|cc|c|c|ccccc|c}
\toprule
\multirow{2}{*}{Method} & \multirow{2}{*}{Type} & \multirow{2}{*}{Segmentor}
& \multirow{2}{*}{\makecell{Params\\(M)}} & \multirow{2}{*}{\makecell{Train.\\(M)}}
& \multirow{2}{*}{\makecell{Src.\\data}} & \multirow{2}{*}{\makecell{Tgt.\\data}}
& \multirow{2}{*}{\makecell{Annotation\\unit}} & \multirow{2}{*}{\makecell{Annot.\\rounds}}
& G2C & S2C & C2A & C2Mu & C2Map & \multirow{2}{*}{Avg.} \\
& & & & & & & &
& \makecell{1/64\\(46)} & \makecell{1/64\\(46)} & \makecell{1/64\\(25)}
& \makecell{1/64\\(23)} & \makecell{1/128\\(140)} & \\
\midrule
\multicolumn{15}{l}{\textit{Full-data references (HRDA/DINOv3-B)}} \\
Target full & Ref. & HRDA/DINOv3-B & 92.31 & 6.64 & -- & \(\checkmark\) & Image & -- &
82.49 & 82.81 & 77.72 & 79.25 & 80.04 & 80.46 \\
Source+target full & Ref. & HRDA/DINOv3-B & 92.31 & 6.64 & \(\checkmark\) & \(\checkmark\) & Image & -- &
81.77 & 82.50 & 78.26 & 80.72 & 80.44 & 80.74 \\
\midrule
\multicolumn{15}{l}{\textit{Native-protocol reproductions}} \\
AllSpark~\cite{allspark} & Semi & SegFormer/MiT-B5 & 84.61 & 84.61 & -- & \(\checkmark\) & Image & 1 &
62.57 & 68.11 & 44.67 & 31.51 & 66.76 & 54.72 \\
UniMatchV2~\cite{unimatchv2} & Semi & DPT/DINOv2-S & 24.80 & 24.80 & -- & \(\checkmark\) & Image & 1 &
72.65 & 73.33 & 57.06 & 51.03 & 69.71 & 64.76 \\
UniMatchV2+Source~\cite{unimatchv2} & SSDA & DPT/DINOv2-S & 24.80 & 24.80 & \(\checkmark\) & \(\checkmark\) & Image & 1 &
75.64 & 79.46 & 69.09 & 67.13 & 76.49 & 73.56 \\
D2ADA~\cite{d2ada} & ADA & DLV2-ASPP256/R101 & 61.38 & 61.38 & \(\checkmark\) & \(\checkmark\) & Region & 5 &
60.84 & 58.68 & 54.26 & 55.02 & 55.47 & 56.85 \\
RIPU~\cite{ripu} & ADA & DLV2-direct/R101 & 43.90 & 43.90 & \(\checkmark\) & \(\checkmark\) & Region & 5 &
66.15 & 68.52 & 57.66 & 57.58 & 64.90 & 62.96 \\
MADAv2~\cite{ning2023madav2} & ADA & DLV3+/R101 & 59.34 & 59.34 & \(\checkmark\) & \(\checkmark\) & Image & 1 &
64.13 & 62.88 & 49.07 & 55.52 & 63.72 & 59.06 \\
HALO~\cite{halo} & ADA & DLV3+-Hyper/R101 & 60.10 & 60.10 & \(\checkmark\) & \(\checkmark\) & Pixel & 5 &
71.87 & 72.63 & 66.39 & 66.70 & 55.65 & 66.65 \\
\midrule
\multicolumn{15}{l}{\textit{Architecture- and annotation-unit-controlled comparison (HRDA/DINOv3-B)}} \\
DIP-HRDA~\cite{dip}\(\dagger\) & FSDA & HRDA/DINOv3-B & 92.31 & 6.64 & \(\checkmark\) & \(\checkmark\) & Image & 1 &
54.48 & 65.18 & 56.83 & 53.35 & 52.40 & 56.45 \\
ADS-SemiSeg+Source~\cite{adssemiseg}\(\dagger\) & SSDA & HRDA/DINOv3-B & 92.31 & 6.64 & \(\checkmark\) & \(\checkmark\) & Image & 1 &
71.30 & 70.67 & 68.82 & 64.02 & 75.72 & 70.11 \\
UniMatchV2+Source~\cite{unimatchv2}\(\dagger\) & SSDA & HRDA/DINOv3-B & 92.31 & 6.64 & \(\checkmark\) & \(\checkmark\) & Image & 1 &
\third{75.86} & \second{78.74} & \second{72.72} & \third{71.25} & \second{77.34} & \second{75.18} \\
D2ADA-Image~\cite{d2ada}\(\dagger\) & ADA & HRDA/DINOv3-B & 92.31 & 6.64 & \(\checkmark\) & \(\checkmark\) & Image & 5 &
68.79 & 69.34 & 65.82 & 69.16 & 70.85 & 68.79 \\
RIPU-Image~\cite{ripu}\(\dagger\) & ADA & HRDA/DINOv3-B & 92.31 & 6.64 & \(\checkmark\) & \(\checkmark\) & Image & 5 &
74.57 & 70.99 & 66.00 & 69.82 & 75.43 & 71.36 \\
MADAv2-HRDA~\cite{ning2023madav2}\(\dagger\) & ADA & HRDA/DINOv3-B & 92.31 & 6.64 & \(\checkmark\) & \(\checkmark\) & Image & 1 &
\second{76.27} & \third{77.18} & \third{70.42} & 70.93 & \third{76.59} & \third{74.28} \\
HALO-Image~\cite{halo}\(\dagger\) & ADA & HRDA/DINOv3-B & 92.31 & 6.64 & \(\checkmark\) & \(\checkmark\) & Image & 5 &
73.00 & 71.80 & 70.30 & \second{71.50} & 75.90 & 72.50 \\
\ours{} & ADA & HRDA/DINOv3-B & 92.31 & 6.64 & \(\checkmark\) & \(\checkmark\) & Image & 1 &
\best{80.64} & \best{81.31} & \best{77.66} & \best{78.15} & \best{79.20} & \best{79.39} \\
\bottomrule
\end{tabular}
\tablenotesep
\parbox{0.98\textwidth}{\footnotesize \emph{Note.} Params and Train. denote the total and trainable segmentor parameters; DLV2/DLV3+ denote DeepLabV2/DeepLabV3+. Dataset abbreviations follow Table~\ref{tab:datasets}. Native-protocol rows retain each method's original segmentor and optimization, whereas \(\dagger\) marks our HRDA/DINOv3-B reproduction. Annotation rounds denote how many times target annotations are collected; semi-supervised baselines and DIP receive one randomly sampled subset before training. ``+Source'' adds source supervision to the original semi-supervised code, with equal numbers of source and target samples in each batch. The top three results in each column of the controlled-comparison block are highlighted. Reference and native-protocol rows are unranked.} 
\vspace{-1em}
\end{table*}

Table~\ref{tab:method_compare} compares methods at the lowest evaluated image-level budget of each transfer: 1/64 for the first four settings and 1/128 for C2Map. Although its ratio is lower, C2Map has a substantially larger target pool, so 1/128 still corresponds to 140 labeled images. Native-protocol reproductions retain each method's original segmentor and optimization, while the final block fixes HRDA/DINOv3-B and the image-level annotation unit to isolate the learning and acquisition strategies. DIP, ADS-SemiSeg, and UniMatchV2 receive a predefined random target subset in one annotation round; the latter two also use the remaining unlabeled images. ``+Source'' means adding source supervision to the corresponding semi-supervised baseline, with source and target samples equally split in each batch. 

Under the controlled segmentor, UniMatchV2+Source is the strongest predefined-subset baseline at 75.18 mIoU on average, while MADAv2 is the strongest one-shot ADA baseline at 74.28. \ours{} reaches 79.39, exceeding them by 4.21 and 5.11 points, respectively, while retaining a single acquisition round. DIP-HRDA and ADS-SemiSeg+Source obtain 56.45 and 70.11, respectively. Unlike \ours{}, neither method actively selects its predefined random subset, and DIP further does not use the remaining unlabeled target pool. \ours{} outperforms the controlled five-round ADA baselines and remains close to the full-data references at the lowest budgets. Despite having 92.31M total parameters, our parameter-efficient configuration optimizes only 6.64M (7.19\%). In the native setting, adding source supervision raises the UniMatchV2 average from 64.76 to 73.56, confirming that source data remains useful when integrated appropriately.

\begin{table}[htbp]
\centering
\caption{Supplementary fixed-count Cityscapes comparison (mIoU, \%).}
\label{tab:fixed_count_results}
\setlength{\tabcolsep}{3.0pt}
\renewcommand{\arraystretch}{1.0}
\begin{tabular}{l|l|cccc}
\toprule
Method & Segmentor & 100 & 200 & 500 & 1000 \\
\midrule
\multicolumn{2}{l}{\textit{GTA $\rightarrow$ Cityscapes-19}} & \multicolumn{4}{r}{\textit{Target full: 82.5}} \\
SemiDAViL~\cite{semidavil}\(\dagger\) & CLIP-B/16+DLG
& 71.1 & 72.5 & 72.9 & 74.8 \\
\ours{} & HRDA/DINOv3-B
& 81.1 & 81.4 & 82.6 & 82.6 \\
Gain & -- & +10.0 & +8.9 & +9.7 & +7.8 \\
\midrule
\multicolumn{2}{l}{\textit{SYNTHIA $\rightarrow$ Cityscapes-16}} & \multicolumn{4}{r}{\textit{Target full: 82.8}} \\
SemiDAViL~\cite{semidavil}\(\dagger\) & CLIP-B/16+DLG
& 76.9 & 77.2 & 78.6 & 79.7 \\
\ours{} & HRDA/DINOv3-B
& 81.7 & 82.1 & 82.9 & 83.3 \\
Gain & -- & +4.8 & +4.9 & +4.3 & +3.6 \\
\bottomrule
\end{tabular}
\tablenotesep
\parbox{0.98\columnwidth}{\footnotesize \emph{Note.} DLG denotes Dense Language Guidance; \(\dagger\) marks values reported in the original paper. Unlike the ratio-based comparisons in the other tables, this table follows SemiDAViL's fixed labeled-image counts. Gain denotes \ours{} minus SemiDAViL in mIoU points.}
\end{table}

SemiDAViL~\cite{semidavil} reports fixed annotation counts rather than ratios. Table~\ref{tab:fixed_count_results} therefore compares the two methods directly at 100/200/500/1000 target images. \ours{} performs better in every setting, with gains of 3.6--10.0 mIoU points.


\subsection{Synthetic-to-Real Results}
Synthetic-to-real adaptation tests one-shot acquisition and target-calibrated adaptation under large appearance shifts. The first two blocks of Table~\ref{tab:ratio_results} compare \ours{} with the controlled baselines across target-label budgets.
The full-data references further show that naively adding synthetic source supervision is not always beneficial: compared with target-only training, joint training changes G2C/S2C by $-0.72/-0.31$ points. This suggests that source supervision can remain detrimental under a large synthetic-to-real gap even when all target labels are available.
At the lowest 1/64 budget (46 images), \ours{} reaches 80.64/81.31 mIoU on G2C/S2C, only 1.85/1.50 points below target-only full supervision. It exceeds the strongest controlled baseline in each setting by 4.37/2.57 points, and the advantage remains consistent as the annotation budget increases.

\vspace{-1.0em}
\subsection{Real-to-Real Results}
The final three blocks of Table~\ref{tab:ratio_results} show the results of adverse-weather and geographically diverse transfers. 
Unlike the synthetic-to-real cases, joint full-data training improves C2A/C2Mu/C2Map by $0.54/1.47/0.40$ points over target-only training, showing that related real source data can remain complementary. UniMatchV2+Source is the strongest controlled baseline across C2A and C2Map, while MADAv2-HRDA and HALO-Image remain competitive on C2Mu. As the budget increases, \ours{} reaches the target-only full-data reference on C2A at 1/16 and surpasses the joint full-data reference at 1/8. On C2Mu, it exceeds target-only training at 1/16, although it remains below joint training. On C2Map, it surpasses both full-data references at 1/32.

\vspace{-1.0em}

\begin{table}[htbp]
\centering
\caption{Image-level comparison across target-label budgets.}  
\label{tab:ratio_results}
\setlength{\tabcolsep}{6.5pt}
\renewcommand{\arraystretch}{1.03}
\begin{tabular}{lcccc}
\toprule
Method & \multicolumn{4}{c}{Target-label budget} \\

\specialrule{0.8pt}{2.2pt}{0pt}
\rowcolor{gray!12}
\multicolumn{5}{c}{\textbf{(a) GTA $\rightarrow$ Cityscapes-19}
\enspace Synthetic-to-real; eval: val} \\
\textit{\makecell{Ratio\\[-1pt](images)}}
& \makecell{1/64\\[-1pt](46)}
& \makecell{1/32\\[-1pt](93)}
& \makecell{1/16\\[-1pt](186)}
& \makecell{1/8\\[-1pt](372)} \\
\cmidrule(lr){1-5}
UniMatchV2~\cite{unimatchv2} & 72.65 & 76.95 & \second{80.17} & \second{80.95} \\
UniMatchV2+Source~\cite{unimatchv2} & \third{75.86} & \second{78.41} & \third{79.17} & \third{80.10} \\
HALO-Image~\cite{halo} & 73.00 & 74.27 & 76.39 & 77.47 \\
MADAv2-HRDA~\cite{ning2023madav2} & \second{76.27} & \third{77.66} & 78.82 & 79.60 \\
\ours{} & \best{80.64} & \best{80.70} & \best{81.24} & \best{82.30} \\
\multicolumn{5}{c}{\textit{References: Target full: 82.49 \quad Source+target full: 81.77}} \\

\specialrule{0.8pt}{2.2pt}{0pt}
\rowcolor{gray!12}
\multicolumn{5}{c}{\textbf{(b) SYNTHIA $\rightarrow$ Cityscapes-16}
\enspace Synthetic-to-real; eval: val} \\
\textit{\makecell{Ratio\\[-1pt](images)}}
& \makecell{1/64\\[-1pt](46)}
& \makecell{1/32\\[-1pt](93)}
& \makecell{1/16\\[-1pt](186)}
& \makecell{1/8\\[-1pt](372)} \\
\cmidrule(lr){1-5}
UniMatchV2~\cite{unimatchv2} & 73.33 & 78.91 & \third{80.61} & \second{81.38} \\
UniMatchV2+Source~\cite{unimatchv2} & \second{78.74} & \second{80.15} & \second{80.65} & 80.79 \\
HALO-Image~\cite{halo} & 71.80 & 74.36 & 76.41 & 77.74 \\
MADAv2-HRDA~\cite{ning2023madav2} & \third{77.18} & \third{78.95} & 79.86 & \third{80.84} \\
\ours{} & \best{81.31} & \best{81.75} & \best{82.15} & \best{82.59} \\
\multicolumn{5}{c}{\textit{References: Target full: 82.81 \quad Source+target full: 82.50}} \\

\specialrule{0.8pt}{2.2pt}{0pt}
\rowcolor{gray!12}
\multicolumn{5}{c}{\textbf{(c) Cityscapes $\rightarrow$ ACDC}
\enspace Real-to-real; eval: val} \\
\textit{\makecell{Ratio\\[-1pt](images)}}
& \makecell{1/64\\[-1pt](25)}
& \makecell{1/32\\[-1pt](50)}
& \makecell{1/16\\[-1pt](100)}
& \makecell{1/8\\[-1pt](200)} \\
\cmidrule(lr){1-5}
UniMatchV2~\cite{unimatchv2} & 57.06 & 64.28 & 64.87 & 69.14 \\
UniMatchV2+Source~\cite{unimatchv2} & \second{72.72} & \second{73.90} & \second{74.27} & \second{75.46} \\
HALO-Image~\cite{halo} & 70.30 & 71.35 & 72.35 & \third{73.98} \\
MADAv2-HRDA~\cite{ning2023madav2} & \third{70.42} & \third{71.96} & \third{72.41} & 73.06 \\
\ours{} & \best{77.66} & \best{77.68} & \best{78.08} & \best{78.50} \\
\multicolumn{5}{c}{\textit{References: Target full: 77.72 \quad Source+target full: 78.26}} \\

\specialrule{0.8pt}{2.2pt}{0pt}
\rowcolor{gray!12}
\multicolumn{5}{c}{\textbf{(d) Cityscapes $\rightarrow$ MUSES}
\enspace Real-to-real; eval: val} \\
\textit{\makecell{Ratio\\[-1pt](images)}}
& \makecell{1/64\\[-1pt](23)}
& \makecell{1/32\\[-1pt](47)}
& \makecell{1/16\\[-1pt](94)}
& \makecell{1/8\\[-1pt](188)} \\
\cmidrule(lr){1-5}
UniMatchV2~\cite{unimatchv2} & 51.03 & 54.09 & 62.70 & 65.27 \\
UniMatchV2+Source~\cite{unimatchv2} & \third{71.25} & \third{72.74} & \third{74.32} & \second{75.45} \\
HALO-Image~\cite{halo} & \second{71.50} & 71.57 & 73.64 & 73.80 \\
MADAv2-HRDA~\cite{ning2023madav2} & 70.93 & \second{72.79} & \second{74.43} & \third{74.83} \\
\ours{} & \best{78.15} & \best{78.64} & \best{79.40} & \best{79.89} \\
\multicolumn{5}{c}{\textit{References: Target full: 79.25 \quad Source+target full: 80.72}} \\

\specialrule{0.8pt}{2.2pt}{0pt}
\rowcolor{gray!12}
\multicolumn{5}{c}{\textbf{(e) Cityscapes $\rightarrow$ Mapillary}
\enspace Real-to-real; eval: val} \\
\textit{\makecell{Ratio\\[-1pt](images)}}
& \makecell{1/128\\[-1pt](140)}
& \makecell{1/64\\[-1pt](281)}
& \makecell{1/32\\[-1pt](563)}
& \makecell{1/16\\[-1pt](1125)} \\
\cmidrule(lr){1-5}
UniMatchV2~\cite{unimatchv2} & 69.71 & 72.05 & 75.72 & 77.46 \\
UniMatchV2+Source~\cite{unimatchv2} & \second{77.34} & \second{78.14} & \second{78.82} & \second{79.40} \\
HALO-Image~\cite{halo} & 75.90 & 76.05 & 76.80 & 77.31 \\
MADAv2-HRDA~\cite{ning2023madav2} & \third{76.59} & \third{77.66} & \third{78.53} & \third{78.79} \\
\ours{} & \best{79.20} & \best{79.92} & \best{80.49} & \best{80.66} \\
\multicolumn{5}{c}{\textit{References: Target full: 80.04 \quad Source+target full: 80.44}} \\
\bottomrule
\end{tabular}
\tablenotesep
\parbox{0.98\columnwidth}{\footnotesize \emph{Note.} Parenthesized values are the numbers of labeled target images. City-19/16 denote the 19-class and shared 16-class protocols. References are unranked. UniMatchV2+Source, HALO-Image, and MADAv2-HRDA use the controlled HRDA/DINOv3-B segmentor; the latter two perform image-level acquisition over five and one rounds, respectively, while UniMatchV2+Source receives one random subset. \ours{} uses a single acquisition--annotation round.}
\vspace{-1.0em}
\end{table}

\subsection{Qualitative Results}

\begin{figure*}[t]
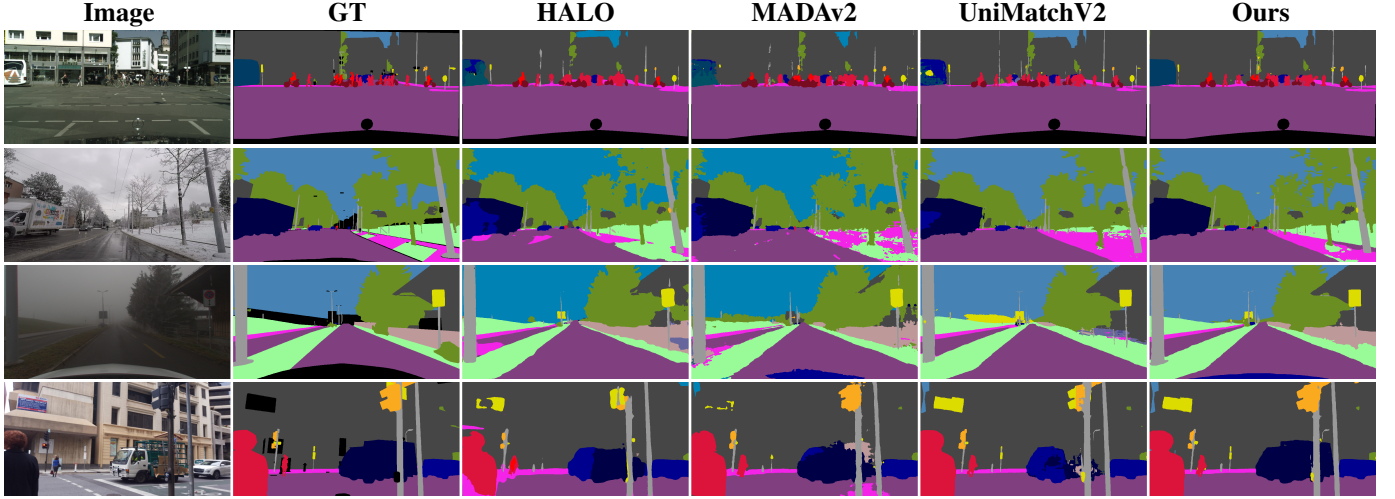

\centering
{\setlength{\tabcolsep}{0pt}%
\begin{tabular}{*{6}{>{\centering\arraybackslash}p{0.1666\textwidth}}}
\textbf{Image} & \textbf{GT} & \textbf{HALO} & \textbf{MADAv2} & \textbf{UniMatchV2} & \textbf{Ours}
\end{tabular}}
\vspace{0.2mm}
\includegraphics[width=\textwidth]{figures/experiment/qualitative_gta2cityscapes.pdf}\par\vspace{0.5mm}
\includegraphics[width=\textwidth]{figures/experiment/qualitative_cityscapes2acdc.pdf}\par\vspace{0.5mm}
\includegraphics[width=\textwidth]{figures/experiment/qualitative_cityscapes2muses.pdf}\par\vspace{0.5mm}
\includegraphics[width=\textwidth]{figures/experiment/qualitative_cityscapes2mapillary.pdf}
\caption{Qualitative comparison at the lowest budgets. From top to bottom: G2C/C2A/C2Mu at 1/64 and C2Map at 1/128. From left to right: input image, ground truth, HALO~\cite{halo}, MADAv2~\cite{ning2023madav2}, UniMatchV2~\cite{unimatchv2}, and \ours{}.
}
\label{fig:qualitative_comparison}
\vspace{-1.5em}
\end{figure*}

Fig.~\ref{fig:qualitative_comparison} shows results at the lowest label budget of each setting. On G2C and C2A, \ours{} better recovers the left-side bus and truck, respectively. On C2A, it also better separates the sidewalk from the large terrain region. On C2Mu, it more accurately segments the right-side fence and the surrounding building and vegetation. On C2Map, it preserves the truck and improves traffic-light and traffic-sign predictions. These examples complement the aggregate results with clearer object integrity and semantic boundaries.
The bottom-connected black areas correspond to ignored ego-vehicle regions in the ground truth.

\vspace{-1.0em}
\subsection{Ablation Study}
We first isolate the representation and sampling choices in Stage~1, then fix the resulting R-DKC selection to study target-calibrated adaptation in Stage~2.

\subsubsection{Stage~1 Ablations}
Stage~1 determines the target representation and ranking. Table~\ref{tab:stage1_embedding_audit} compares layouts with R-DKC, Table~\ref{tab:stage1_selection_ablation} compares samplers with G2x4, and Tables~\ref{tab:stage1_evidence_audit} and~\ref{tab:rare_component_ablation} analyze R. All rows use the same Stage~2 setting adopted in the main experiments, so only the Stage~1 representation or ranking changes. We then fix G2x4 with R-DKC and ablate Stage~2.

\begin{table}[htbp]
\centering
\caption{Embedding-layout ablation on C2A with sampling fixed (mIoU, \%).} %
\label{tab:stage1_embedding_audit}
\renewcommand{\arraystretch}{1.0}
\begin{tabular}{lc|ccc|r}
\toprule
Layout & CN & 1/64 & 1/16 & Avg. & \makecell{Storage $\downarrow$\\(MiB)} \\
\midrule
Global & -- & 76.95 & 77.49 & 77.22 & \best{19} \\
$+$G2x2 & yes & 77.19 & \second{78.04} & \third{77.62} & \second{94} \\
$+$G4x4 & yes & \third{77.37} & 77.79 & 77.58 & 319 \\
$+$G4x8 & yes & 76.93 & \third{78.02} & 77.47 & 619 \\
$+$G8x8 & yes & \second{77.58} & 77.83 & \second{77.70} & 1219 \\
\textbf{$+$G2x4} & yes & \best{77.66} & \best{78.08} & \best{77.87} & \third{169} \\
\bottomrule
\end{tabular}
\tablenotesep
\parbox{\columnwidth}{\footnotesize \emph{Note.} Every grid includes the global descriptor. G$h{\times}w$ denotes an $h\times w$ grid. CN denotes component normalization. Storage is the cached embedding size in MiB and is ranked in ascending order. All rows use the same four-layer VFM, R-DKC, and Stage~2 setting.
}
\vspace{-1.0em}
\end{table}

\begin{table}[htbp]
\centering
\caption{Comparison of Stage~1 sampling strategies (mIoU, \%).}
\label{tab:stage1_selection_ablation}
\renewcommand{\arraystretch}{0.98}
\begin{tabular}{l|ccc|c}
\toprule
\multirow{2}{*}{Sampler} & \multicolumn{3}{c|}{C2A} & C2Map \\
& 1/64 & 1/16 & Avg. & 1/128 \\
\midrule
Random & 76.58 & 77.56 & 77.07 & 78.75 \\
Uniform & 77.14 & \third{77.83} & \third{77.48} & 79.01 \\
\midrule
Entropy & 76.65 & 77.56 & 77.11 & 78.54 \\
K-center~\cite{k-center} & \third{77.27} & 77.57 & 77.42 & 78.09 \\
Density-k-center~\cite{ssmis} & 77.17 & 77.69 & 77.43 & 78.90 \\
K-Medoids & 77.15 & 77.78 & 77.47 & \second{79.09} \\
TypiClust~\cite{typiclust} & 76.68 & 77.41 & 77.05 & \third{79.06} \\
BADGE~\cite{badge} & \second{77.40} & 77.59 & \second{77.50} & 78.96 \\
\midrule
Pseudo-class-only & 76.96 & \second{77.93} & 77.45 & 79.03 \\
R-DKC & \best{77.66} & \best{78.08} & \best{77.87} & \best{79.20} \\
\bottomrule
\end{tabular}
\tablenotesep
\parbox{\columnwidth}{\footnotesize \emph{Note.} Random and Uniform require no embedding; all feature-based methods use G2x4. Every row shares the same Stage~2 setting. Pseudo-class-only ranks images using R without density--diversity coverage, whereas R-DKC combines R with DKC. } 
\vspace{-1.0em}
\end{table}

\textbf{Embedding representation.} Table~\ref{tab:stage1_embedding_audit} compares global and grid representations under the R evidence. Global$+$G2x4 achieves the highest mean accuracy across the two budgets and is therefore adopted as the default representation.

\textbf{Sampling strategy.}
Table~\ref{tab:stage1_selection_ablation} compares random and uniform selection with representative diversity, clustering, uncertainty, coverage, and semantic acquisition rules. Feature-based methods use the final G2x4 representation, and every row shares the same target-calibrated adaptation configuration, so differences arise only from target selection.

\begin{table}[htbp]
\centering
\caption{Semantic coverage and feature diversity of selected ACDC subsets.}
\label{tab:stage1_split_metrics}
\footnotesize
\setlength{\tabcolsep}{1.6pt}
\renewcommand{\arraystretch}{1.03}
\begin{tabular}{l|cc|cc|cc}
\toprule
\multirow{2}{*}{Selection}
& \multicolumn{2}{c|}{Class entropy $\uparrow$}
& \multicolumn{2}{c|}{Rare enrich. $\uparrow$}
& \multicolumn{2}{c}{Diversity $\uparrow$} \\
& 1/64 & 1/16 & 1/64 & 1/16 & 1/64 & 1/16 \\
\midrule
Random & 1.8351 & 1.8569 & 1.43 & 1.10 & 0.1352 & 0.1369 \\
Uniform & 1.8932 & 1.9046 & 1.93 & 1.44 & 0.1220 & 0.1432 \\ \midrule
Entropy & 1.6971 & 1.7169 & 0.69 & 0.88 & 0.1309 & 0.1257 \\
K-center~\cite{k-center} & \third{1.9846} & \third{1.9957} & \third{3.11} & \third{2.42} & \second{0.1715} & \best{0.1546} \\
Density-k-center~\cite{ssmis} & 1.8816 & 1.9695 & 1.67 & 2.07 & \best{0.1731} & \second{0.1528} \\
K-Medoids & 1.8265 & 1.9190 & 0.81 & 1.47 & 0.1291 & 0.1361 \\
TypiClust~\cite{typiclust} & 1.7910 & 1.8426 & 0.58 & 0.62 & 0.1388 & 0.1314 \\
BADGE~\cite{badge} & 1.9784 & 1.9331 & 2.21 & 1.61 & 0.1541 & 0.1426 \\ \midrule
Pseudo-class-only & \second{2.0634} & \best{2.1062} & \best{5.85} & \best{5.29} & 0.1383 & 0.1409 \\
R-DKC & \best{2.1138} & \second{2.0597} & \second{5.22} & \second{3.49} & \third{0.1706} & \third{0.1512} \\
\bottomrule
\end{tabular}
\tablenotesep
\parbox{\columnwidth}{\footnotesize \emph{Note.} Rankings are assigned separately within each metric and budget. Target ground truth is used only for post-hoc analysis and never affects acquisition. All feature-based methods use the same G2x4 representation.}
\vspace{-1.0em}
\end{table}

To examine what each acquisition rule selects beyond downstream mIoU, Table~\ref{tab:stage1_split_metrics} evaluates the same methods as Table~\ref{tab:stage1_selection_ablation} using the semantic coverage and feature redundancy of their ACDC subsets at the 1/64 and 1/16 budgets. Class entropy is $-\sum_c p_c\ln p_c$, where $p_c$ is the class-$c$ pixel proportion within a selected subset and a larger value indicates more balanced semantic coverage. Rare-class enrichment divides the selected subset's rare-class pixel share by its full-pool share, so values above one indicate increased coverage. Rare classes have a nonzero full-pool pixel frequency below $0.5\%$; on ACDC, they are traffic light, traffic sign, person, rider, truck, bus, train, motorcycle, and bicycle. Diversity is the mean pairwise cosine distance between selected embeddings, with larger values indicating less redundancy.



\textbf{Semantic evidence.} We add each candidate semantic cue separately to DKC to identify which one transfers reliably across datasets.
Table~\ref{tab:stage1_evidence_audit} shows that rare-class coverage is the only cue that improves the dataset-level average over DKC on all four transfers, yielding the largest gain of 0.52 points. Uncertainty, entropy, and confidence help in isolated settings but are inconsistent across domains. We therefore retain rare-class coverage as the semantic evidence in the final selector.

\begin{table}[htbp]
\centering
\caption{Individual semantic cues added separately to DKC with fixed G2x4 embeddings (mIoU, \%).}
\label{tab:stage1_evidence_audit}
\setlength{\tabcolsep}{2.8pt}
\renewcommand{\arraystretch}{1.0}
\begin{tabular}{l|cc|cc|c|c|c}
\toprule
\multirow{2}{*}{Evidence} & \multicolumn{2}{c|}{C2A} & \multicolumn{2}{c|}{G2C} & C2Map & C2Mu & \multirow{2}{*}{$\Delta$DKC} \\
& 1/64 & 1/16 & 1/64 & 1/16 & 1/128 & 1/16 & \\
\midrule
None (DKC) & \third{77.17} & \third{77.69} & \second{80.21} & 81.20 & \third{78.90} & 78.28 & 0.00 \\
Uncertainty & 77.13 & \best{78.46} & 79.76 & \third{81.25} & 78.62 & 77.90 & -0.13 \\
Entropy & \second{77.62} & 77.19 & 79.99 & \best{81.66} & 78.84 & \third{78.29} & \third{+0.01} \\
Confidence & 76.98 & 77.13 & \third{80.17} & \second{81.33} & \best{79.29} & \second{78.49} & \second{+0.07} \\
\midrule
\textbf{Rare coverage (R)} & \best{77.66} & \second{{78.08}} & \best{80.64} & {81.24} & \second{{79.20}} & \best{79.40} & \best{+0.52} \\
\bottomrule
\end{tabular}
\tablenotesep
\parbox{\columnwidth}{\footnotesize \emph{Note.} R denotes confidence-weighted rare-class coverage, and the bold row is selected for the final method. All rows share the G2x4 embeddings and final transfer-specific Stage~2 profiles. $\Delta$DKC averages each dataset internally before averaging the four dataset-level differences.}
\vspace{-1.5em}
\end{table}

\begin{table}[htbp]
\centering
\caption{Internal components of reliable rare-class coverage on C2A (mIoU, \%).}
\label{tab:rare_component_ablation}
\setlength{\tabcolsep}{2.5pt}
\renewcommand{\arraystretch}{1.0}
\begin{tabular}{l|ccc|ccc}
\toprule
Configuration & Area & Conf. & Rarity & 1/64 & 1/16 & Avg. \\
\midrule
DKC baseline & -- & -- & -- & 77.17 & 77.69 & 77.43 \\
w/o area scaling & -- & \cmark & \cmark & \third{77.46} & 77.67 & \third{77.57} \\
w/o class confidence & \cmark & -- & \cmark & 76.78 & \second{77.93} & 77.36 \\
w/o rarity weighting & \cmark & \cmark & -- & \best{77.73} & \third{77.82} & \second{77.78} \\
\midrule
\textbf{Full R} & \cmark & \cmark & \cmark & \second{77.66} & \best{78.08} & \best{77.87} \\
\bottomrule
\end{tabular}
\tablenotesep
\parbox{\columnwidth}{\footnotesize \emph{Note.} Area denotes square-root area scaling, Conf. the class-region mean confidence, and Rarity the inverse-frequency class weight. The area threshold that removes tiny pseudo-label fragments is retained in every R variant. All rows use the final C2A Stage~2 profile.}
\vspace{-1.0em}
\end{table}



\begin{table*}[htbp]
\centering
\caption{ADA accuracy (mIoU, \%) and acquisition cost under matched image-level budgets.}
\label{tab:acquisition_protocol_cost}
\renewcommand{\arraystretch}{1.02}
\begin{tabular}{lcc|rr|rr|rr|rr}
\toprule
Method & Mode & Rounds & \multicolumn{2}{c|}{G2C 1/64} & \multicolumn{2}{c|}{C2A 1/64} & \multicolumn{2}{c|}{C2Map 1/128} & \multicolumn{2}{c}{Avg.} \\
& & & mIoU & min & mIoU & min & mIoU & min & mIoU & min \\
\midrule
RIPU-Image~\cite{ripu} & Online & 5 & \third{74.57} & \third{93.7} & 66.00 & 53.5 & 75.43 & 738.0 & 72.00 & 295.1 \\
D2ADA-Image~\cite{d2ada} & Online & 5 & 68.79 & 1095.5 & 65.82 & 232.4 & 70.85 & \third{381.9} & 68.49 & 569.9 \\
HALO-Image~\cite{halo} & Online & 5 & 73.00 & \second{70.2} & \third{70.30} & \third{36.3} & \third{75.90} & 411.0 & \third{73.07} & \third{172.5} \\
MADAv2-HRDA~\cite{ning2023madav2} & Offline & 1 & \second{76.27} & 165.4 & \second{70.42} & \second{26.2} & \second{76.59} & \second{138.9} & \second{74.43} & \second{110.2} \\ 
\ours{} (R-DKC) & Offline & 1 & \best{80.64} & \best{17.7} & \best{77.66} & \best{8.9} & \best{79.20} & \best{107.0} & \best{79.17} & \best{44.5} \\
\bottomrule
\end{tabular}
\tablenotesep
\parbox{0.98\textwidth}{\footnotesize \emph{Note.} Acquisition times are measured on a single RTX 4090 and include pool inference and selection over all acquisition rounds; segmentation training and human annotation are excluded. All rows use complete-image budgets and HRDA/DINOv3-B. External methods retain their released training loops.} 
\vspace{-1.0em}
\end{table*}

\begin{table*}[ht]
\centering
\caption{Ablation of target-calibrated adaptation on C2A and G2C (mIoU, \%).}
\label{tab:ablation}
\renewcommand{\arraystretch}{1.0}
\begin{tabular}{l|cccccc|c|cc}
\toprule
\multirow{2}{*}{Setting} & \multicolumn{6}{c|}{Stage~2 design} &
\multirow{2}{*}{Init.} & C2A & G2C \\
& $S\!\to\!T_u$ & $S\!\to\!T_l$ & T-Sup & T-Mix & Atten. & TDM &
& 1/64 & 1/64 \\
\midrule
Target full supervision & -- & -- & -- & -- & -- & -- & -- &
77.72 & 82.49 \\
Source only & -- & -- & -- & -- & -- & -- & -- &
70.32 & 69.79 \\
\midrule
DACS ($S\!\to\!T_u$) & \cmark & -- & -- & -- & -- & -- & scratch &
72.92 & 77.13 \\
$+$ labeled-target context & \cmark & \cmark & -- & -- & -- & -- & scratch &
76.56 & 79.37 \\
$+$ direct T-Sup & \cmark & \cmark & \cmark & -- & -- & -- & scratch &
75.87 & 78.60 \\
$+$ attenuation & \cmark & \cmark & \cmark & -- & \cmark & -- & scratch &
76.17 & 78.81 \\
$+$ T-Mix & \cmark & \cmark & -- & \cmark & -- & -- & scratch &
76.70 & 80.05 \\
Four-branch objective & \cmark & \cmark & \cmark & \cmark & -- & -- & scratch &
76.89 & 79.71 \\
\midrule
$+$ TDM & \cmark & \cmark & \cmark & \cmark & -- & \cmark & scratch &
76.93 & 79.85 \\
$+$ UDA initialization & \cmark & \cmark & \cmark & \cmark & -- & -- & UDA &
\third{77.45} & \third{80.44} \\
$+$ TDM $+$ UDA initialization & \cmark & \cmark & \cmark & \cmark & -- & \cmark & UDA &
\second{77.54} & \second{80.57} \\
\textbf{$+$ attenuation} & \cmark & \cmark & \cmark & \cmark & \cmark & \cmark & UDA &
\best{77.66} & \best{80.64} \\
\bottomrule
\end{tabular}
\tablenotesep
\parbox{0.98\textwidth}{\footnotesize \emph{Note.} Source supervision is active in every Stage~2 ablation row. The two reference rows use no Stage~1 acquisition. DACS mixes source images only with unlabeled target contexts; the hybrid branch additionally uses labeled target contexts. Scratch denotes random adapter/decoder initialization rather than random target selection. Atten. denotes budget-aware T-Sup attenuation. UDA initialization restores only the adapters and decoder and resets all optimization states.}
\vspace{-1.5em}
\end{table*}

\begin{table}[htbp]
\centering
\caption{Breakdown of our one-shot acquisition time (minutes).}
\label{tab:stage1_cost}
\setlength{\tabcolsep}{3.5pt}
\renewcommand{\arraystretch}{1.02}
\begin{tabular}{@{}lc|rrrr@{}}
\toprule
Transfer & Budget & \makecell{VFM\\encode} & \makecell{UDA\\stats}
& \makecell{R-DKC\\ranking} & Total \\
\midrule
G2C & 1/64 & $2.93{\pm}0.07$ & $14.57{\pm}0.02$ & $0.23{\pm}0.00$ & $17.72{\pm}0.09$ \\
C2A & 1/64 & $1.21{\pm}0.16$ & $7.53{\pm}0.04$ & $0.13{\pm}0.01$ & $8.87{\pm}0.19$ \\
C2Map & 1/128 & $20.70{\pm}0.32$ & $83.14{\pm}0.40$ & $3.19{\pm}0.11$ & $107.02{\pm}0.74$ \\
\bottomrule
\end{tabular}
\tablenotesep
\parbox{\columnwidth}{\footnotesize \emph{Note.} Values are three-run means $\pm$ sample standard deviations on one RTX 4090. Total includes VFM encoding, one fixed-teacher pass for UDA statistics, and R-DKC ranking, and is computed from unrounded times.}
\end{table}

\begin{table}[t]
\centering
\caption{Three-run reliability at the lowest budget (mIoU, \%; mean $\pm$ sample standard deviation).}
\label{tab:low_budget_reliability}
\setlength{\tabcolsep}{4pt}
\renewcommand{\arraystretch}{1.02}
\begin{tabular}{l|l|cc}
\toprule
Method & Varied factor & G2C 1/64 & C2A 1/64 \\
\midrule
Random & Selection seed & $80.07\pm0.36$ & $76.86\pm0.33$ \\
\ours{} & Training seed & $80.55\pm0.11$ & $77.67\pm0.09$ \\
\bottomrule
\end{tabular}
\tablenotesep
\parbox{0.98\columnwidth}{\footnotesize \emph{Note.} Random fixes training seed 0 and varies three independently sampled target subsets. \ours{} fixes its deterministic R-DKC subset and varies the training seed. Both use the same Stage~2 configuration and checkpoint rule.}
\vspace{-1em}
\end{table}

\begin{table}[t]
\centering
\caption{Progressive acquisition on synthetic-to-real transfers (mIoU, \%).}
\label{tab:progressive_extension}
\footnotesize
\setlength{\tabcolsep}{3pt}
\renewcommand{\arraystretch}{1.02}
\begin{tabular}{l|ccc|ccc}
\toprule
& \multicolumn{3}{c|}{G2C} & \multicolumn{3}{c}{S2C} \\
\cmidrule(lr){2-4}\cmidrule(lr){5-7}
Setting & 1/64 & 1/32 & 1/16 & 1/64 & 1/32 & 1/16 \\
\midrule
One-shot & 80.64 & 80.70 & 81.24 & 81.31 & 81.75 & 82.15 \\
Progressive & 80.49 & 81.25 & 81.67 & 81.38 & 81.91 & 82.36 \\
$\Delta$ & -0.15 & +0.55 & +0.43 & +0.07 & +0.16 & +0.21 \\ \midrule
Avg. extra time (min) & \multicolumn{3}{c|}{23.2} & \multicolumn{3}{c}{27.7} \\
\bottomrule
\end{tabular}
\tablenotesep
\parbox{\columnwidth}{\footnotesize \emph{Note.} One-shot annotates all $K$ images once; progressive uses two rounds of $K/2$ images. Average extra time covers the second online inference and selection round over the three budgets of each dataset.} 
\vspace{-1em}
\end{table}

\begin{table}[t]
\centering
\caption{Generalization across segmentation models (mIoU, \%).}
\label{tab:model_expansion}
\setlength{\tabcolsep}{4pt}
\renewcommand{\arraystretch}{1.06}
\begin{tabular}{l|ccccc}
\toprule
Model & Full sup & Sup@r & Semi@r & \ours{}@r & $+$KD \\
\midrule
\multicolumn{6}{l}{\textit{GTA $\rightarrow$ Cityscapes; $r=1/8$; target eval: val}} \\
DINOv3-L & 83.88 & 81.88 & 83.39 & 83.55 & -- \\
DINOv2-B & 82.47 & 79.88 & 81.54 & 82.15 & -- \\
PIDNet-S & 75.59 & 67.74 & 71.96 & 72.80 & 74.45 \\
\midrule
\multicolumn{6}{l}{\textit{Cityscapes $\rightarrow$ ACDC; $r=1/16$; target eval: val}} \\
DINOv3-L & 80.27 & 76.34 & 78.73 & 80.44 & -- \\
DINOv2-B & 78.16 & 70.81 & 74.43 & 77.97 & -- \\
PIDNet-S & 64.55 & 51.79 & 55.79 & 63.49 & 66.79 \\
\bottomrule
\end{tabular}
\tablenotesep
\parbox{0.98\columnwidth}{\footnotesize \emph{Note.} Sup@r and Semi@r denote target-only supervised and semi-supervised training at ratio $r$, respectively; TC-ADA@r uses the same selected target subset. DINOv3-L and DINOv2-B use Rein and HRDA. +KD adds offline DINOv3-L pseudo labels without additional target annotation.}
\vspace{-2.0em}
\end{table}

Table~\ref{tab:rare_component_ablation} further isolates the three factors within R. Removing class confidence lowers the average from 77.87 to 77.36, including a drop from 77.66 to 76.78 at 1/64. This suggests that suppressing unreliable rare-class predictions is especially important when only 25 target images can be annotated. Removing area scaling and rarity weighting gives averages of 77.57 and 77.78, respectively. Although the variant without rarity weighting is strongest at 1/64, the complete R score improves the DKC average from 77.43 to 77.87 and achieves the best cross-budget result. 

\subsubsection{Acquisition Efficiency}
Tables~\ref{tab:stage1_cost} and~\ref{tab:acquisition_protocol_cost} provide complementary efficiency analyses. The former decomposes the one-shot cost of R-DKC, whereas the latter compares complete ADA methods by accuracy and end-to-end acquisition time.

Table~\ref{tab:stage1_cost} shows that R-DKC ranking takes only 0.13--3.19 minutes and most acquisition time comes from the one-time VFM encoding and fixed-teacher statistics. Table~\ref{tab:acquisition_protocol_cost} provides the broader protocol comparison. Iterative methods repeatedly infer over the target pool and interrupt adaptation for new annotations, whereas \ours{} generates its ranking once. Compared with one-shot MADAv2, TC-ADA improves the three-transfer average by 4.74 points (79.17 vs.\ 74.43) and reduces mean acquisition time from 110.2 to 44.5 minutes.

\subsubsection{Stage~2 Objective and Calibration}

Table~\ref{tab:ablation} first provides the target-domain upper reference and the source-only starting point. Starting from DACS, adding labeled-target contexts improves the 1/64 result by 3.64/2.24 points on C2A/G2C. Direct T-Sup alone is less effective; attenuating it at low budgets recovers 0.30/0.21 points under the same random initialization. T-Mix further improves the hybrid objective, whereas adding unit-weight direct T-Sup to form the four-branch objective changes C2A/G2C by $+0.19/-0.34$ points. This dataset-dependent effect motivates budget-aware attenuation rather than treating direct target supervision as uniformly reliable.

TDM alone gives smaller gains of 0.04/0.14 points, while UDA initialization improves C2A/G2C by 0.56/0.73 points. Under the same UDA initialization, adding TDM reaches 77.54/80.57 mIoU; attenuation further improves the paired results to 77.66/80.64. These results support attenuation as the default budget-aware weighting for direct target supervision. 

\subsubsection{Low-Budget Statistical Reliability}

Table~\ref{tab:low_budget_reliability} separates target-subset variability from optimization variability at the lowest budgets. Across three runs, \ours{} improves the Random mean by 0.48/0.81 points on G2C/C2A. Its fixed R-DKC subset also yields small training deviations of 0.11/0.09, whereas independently sampled Random subsets vary by 0.36/0.33. Because the rows vary different sources of randomness, these deviations are diagnostic rather than a paired significance test; nevertheless, the reported comparison is not based on a single optimization seed or a single Random subset.

\subsection{Optional Progressive Acquisition}

The default \ours{} remains one-shot. As an optional two-round extension, it first acquires $K/2$ images, then uses the EMA model after a 4k-iteration warm-up to select another $K/2$ from the remaining pool. Table~\ref{tab:progressive_extension} reports accuracy and the extra acquisition time. 
Progressive acquisition improves five of six settings by 0.07--0.55 points, but decreases G2C at 1/64 by 0.15 points and adds 23.2/27.7 minutes on G2C/S2C. Given these variable gains and extra cost, we retain one-shot acquisition as the default experiment setting.

\subsection{Model Generalization}
We evaluate \ours{} on different segmentation backbones.
Table~\ref{tab:model_expansion} reports results with DINOv3-L, DINOv2-B, and real-time PIDNet-S~\cite{pidnet} on G2C and C2A.
Each VFM constructs its own R-DKC split, while PIDNet-S reuses the DINOv3-L split. The PIDNet-S distillation results use offline pseudo labels produced only after the corresponding DINOv3-L \ours{} teacher has finished training, without changing the real-time inference architecture.

Across the two VFM segmentors, \ours{} remains within 0.33 points of full supervision on G2C. On C2A, it exceeds the DINOv3-L reference by 0.17 points and is only 0.19 points below the DINOv2-B reference. For PIDNet-S, \ours{} improves target-only semi-supervision by 0.84/7.70 points on G2C/C2A. Offline distillation adds another 1.65/3.30 points, narrowing the G2C gap and surpassing the C2A full-supervision reference while retaining the real-time model.

\FloatBarrier

\section{Conclusion}
In this paper, we presented TC-ADA, a systematic joint design under the one-shot image-level ADA setting for label-efficient semantic segmentation in driving scenes. Stage~1 selects representative and informative target images in a single round using VFM-based visual representations and UDA-derived semantic information. Stage~2 jointly uses labeled source, labeled target, and unlabeled target data, while calibrating source and target supervision under limited target labels.

Experiments across five driving transfers show consistent improvements over representative ADA baselines under controlled image-level settings. With only 1/64 of the target images labeled on four transfers and 1/128 on the Mapillary transfer, TC-ADA remains within 1.9 mIoU points of target-only full supervision on all five transfers. Ablation studies further support the effectiveness of both target selection and target-calibrated adaptation. Future work will explore the extension of TC-ADA to other vision tasks and more challenging target-domain shifts.

\bibliographystyle{IEEEtran}
\bibliography{IEEEabrv,reference.bib}

@ARTICLE{semi_survey,
  author={Ran, Lingyan and Li, Yali and Liang, Guoqiang and Zhang, Yanning},
  journal={IEEE Transactions on Circuits and Systems for Video Technology}, 
  title={Pseudo Labeling Methods for Semi-Supervised Semantic Segmentation: A Review and Future Perspectives}, 
  year={2025},
  volume={35},
  number={4},
  pages={3054-3080},
  doi={10.1109/TCSVT.2024.3508768}
}

@inproceedings{cityscapes,
  title={The {Cityscapes} Dataset for Semantic Urban Scene Understanding},
  author={Cordts, Marius and Omran, Mohamed and Ramos, Sebastian and Rehfeld, Timo and Enzweiler, Markus and Benenson, Rodrigo and Franke, Uwe and Roth, Stefan and Schiele, Bernt},
  booktitle={Proceedings of the IEEE Conference on Computer Vision and Pattern Recognition},
  pages={3213--3223},
  year={2016}
}

@inproceedings{playingfordata,
  title={Playing for Data: Ground Truth from Computer Games},
  author={Richter, Stephan R. and Vineet, Vibhav and Roth, Stefan and Koltun, Vladlen},
  booktitle={European Conference on Computer Vision},
  pages={102--118},
  year={2016}
}

@inproceedings{synthia,
  title={The {SYNTHIA} Dataset: A Large Collection of Synthetic Images for Semantic Segmentation of Urban Scenes},
  author={Ros, German and Sellart, Laura and Materzynska, Joanna and Vazquez, David and Lopez, Antonio M.},
  booktitle={Proceedings of the IEEE Conference on Computer Vision and Pattern Recognition},
  pages={3234--3243},
  year={2016}
}

@inproceedings{acdc,
  title={{ACDC}: The Adverse Conditions Dataset with Correspondences for Semantic Driving Scene Understanding},
  author={Sakaridis, Christos and Dai, Dengxin and Van Gool, Luc},
  booktitle={Proceedings of the IEEE/CVF International Conference on Computer Vision},
  pages={10765--10775},
  year={2021}
}

@inproceedings{muses,
  title={{MUSES}: The Multi-Sensor Semantic Perception Dataset for Driving under Uncertainty},
  author={Br{\"o}dermann, Tim and Bruggemann, David and Sakaridis, Christos and Ta, Kevin and Liagouris, Odysseas and Corkill, Jason and Van Gool, Luc},
  booktitle={European Conference on Computer Vision},
  pages={21--38},
  year={2024}
}

@inproceedings{mapillary,
  title={The {Mapillary Vistas} Dataset for Semantic Understanding of Street Scenes},
  author={Neuhold, Gerhard and Ollmann, Tobias and Rota Bulo, Samuel and Kontschieder, Peter},
  booktitle={Proceedings of the IEEE International Conference on Computer Vision},
  pages={4990--4999},
  year={2017}
}

@article{CAM,
  title={Context-Aware Mixup for Domain Adaptive Semantic Segmentation},
  author={Zhou, Qianyu and Feng, Zhengyang and Gu, Qiqi and Pang, Jiangmiao and Cheng, Guangliang and Lu, Xuequan and Shi, Jianping and Ma, Lizhuang},
  journal={IEEE Transactions on Circuits and Systems for Video Technology},
  year={2023},
  volume={33},
  number={2},
  pages={804--817},
  doi={10.1109/TCSVT.2022.3206476},
  publisher={IEEE}
}

@ARTICLE{scsd,
  author={Qin, Jinghui and Tan, Hao and Jin, Kebing},
  journal={IEEE Transactions on Circuits and Systems for Video Technology}, 
  title={One-shot Unsupervised Domain Adaptation for Semantic Segmentation with Self-Constrained Style Decoupling}, 
  year={2026},
  note={early access, doi: 10.1109/TCSVT.2026.3702681},
  doi={10.1109/TCSVT.2026.3702681}
}

@article{tufl,
  author={Yan, Weihao and Qian, Yeqiang and Wang, Chunxiang and Yang, Ming},
  journal={IEEE Transactions on Intelligent Transportation Systems}, 
  title={Threshold-Adaptive Unsupervised Focal Loss for Domain Adaptation of Semantic Segmentation}, 
  year={2023},
  volume={24},
  number={1},
  pages={752--763},
}

@article{sam4udass,
  author={Yan, Weihao and Qian, Yeqiang and Zhuang, Hanyang and Wang, Chunxiang and Yang, Ming},
  journal={IEEE Transactions on Intelligent Vehicles}, 
  title={{SAM4UDASS}: When {SAM} Meets Unsupervised Domain Adaptive Semantic Segmentation in Intelligent Vehicles}, 
  year={2024},
  volume={9},
  number={2},
  pages={3396--3408},
}

@inproceedings{dacs,
  title={{DACS}: Domain Adaptation via Cross-Domain Mixed Sampling},
  author={Tranheden, Wilhelm and Olsson, Viktor and Pinto, Juliano and Svensson, Lennart},
  booktitle={Proceedings of the IEEE/CVF Winter Conference on Applications of Computer Vision},
  pages={1379--1389},
  year={2021}
}

@inproceedings{daformer,
  title={{DAFormer}: Improving Network Architectures and Training Strategies for Domain-Adaptive Semantic Segmentation},
  author={Hoyer, Lukas and Dai, Dengxin and Van Gool, Luc},
  booktitle={Proceedings of the IEEE/CVF Conference on Computer Vision and Pattern Recognition},
  pages={9924--9935},
  year={2022}
}

@inproceedings{hrda,
  title={{HRDA}: Context-Aware High-Resolution Domain-Adaptive Semantic Segmentation},
  author={Hoyer, Lukas and Dai, Dengxin and Van Gool, Luc},
  booktitle={European Conference on Computer Vision},
  pages={372--391},
  year={2022}
}

@inproceedings{mic,
  title={{MIC}: Masked Image Consistency for Context-Enhanced Domain Adaptation},
  author={Hoyer, Lukas and Dai, Dengxin and Wang, Haoran and Van Gool, Luc},
  booktitle={Proceedings of the IEEE/CVF Conference on Computer Vision and Pattern Recognition},
  pages={11721--11732},
  year={2023}
}

@article{vfmuda,
  title={Rethinking Domain Adaptive Semantic Segmentation With Vision Foundation Models},
  author={Liao, Muxin and Peng, Yingqiong and Sun, Yuting and Huang, Wenju and Wang, Yinglong and Zhang, Yuhang},
  journal={IEEE Transactions on Multimedia},
  year={2026},
  pages={1--14},
  doi={10.1109/TMM.2026.3687760},
  publisher={IEEE}
}

@inproceedings{rein,
  title={Stronger Fewer and Superior: Harnessing Vision Foundation Models for Domain Generalized Semantic Segmentation},
  author={Wei, Zhixiang and Chen, Lin and Jin, Yi and Ma, Xiaoxiao and Liu, Tianle and Ling, Pengyang and Wang, Ben and Chen, Huaian and Zheng, Jinjin},
  booktitle={Proceedings of the IEEE/CVF Conference on Computer Vision and Pattern Recognition},
  pages={28619--28630},
  year={2024}
}

@article{reinpp,
  author={Wei, Zhixiang and Ma, Xiaoxiao and Yan, Ruishen and Tu, Tao and Chen, Huaian and Zheng, Jinjin and Jin, Yi and Chen, Enhong},
  journal={IEEE Transactions on Pattern Analysis and Machine Intelligence}, 
  title={{Rein++}: Efficient Generalization and Adaptation for Semantic Segmentation with Vision Foundation Models},
  year={2026},
  volume={48},
  number={8},
  pages={9826--9843},
  doi={10.1109/TPAMI.2026.3681631}}

@article{dinov2,
  title={{DINOv2}: Learning Robust Visual Features without Supervision},
  author={Oquab, Maxime and Darcet, Timoth{\'e}e and Moutakanni, Th{\'e}o and Vo, Huy V. and Szafraniec, Marc and Khalidov, Vasil and Fernandez, Pierre and Haziza, Daniel and Massa, Francisco and others},
  journal={Transactions on Machine Learning Research},
  year={2024}
}

@article{dinov3,
  title={{DINOv3}},
  author={Sim{\'e}oni, Oriane and Vo, Huy V. and Seitzer, Maximilian and Baldassarre, Federico and Oquab, Maxime and Jose, Cijo and Khalidov, Vasil and Szafraniec, Marc and Yi, Seungeun and Ramamonjisoa, Micha{\"e}l and others},
  journal={arXiv preprint arXiv:2508.10104},
  year={2025}
}

@article{adaptiveseg,
  title={An adaptive post-processing network with the global-local aggregation for semantic segmentation},
  author={Zhu, Guilin and Wang, Runmin and Liu, Yingying and Zhu, Zhenlin and Gao, Changxin and Liu, Li and Sang, Nong},
  journal={IEEE Transactions on Circuits and Systems for Video Technology},
  year={2024},
  volume={34},
  number={2},
  pages={1159--1173},
  publisher={IEEE}
}

@article{infor_tcsvt,
  author={Wu, Jiawei and Fan, Haoyi and Li, Zuoyong and Liu, Guang-Hai and Lin, Shouying},
  journal={IEEE Transactions on Circuits and Systems for Video Technology}, 
  title={Information Transfer in Semi-Supervised Semantic Segmentation}, 
  year={2024},
  volume={34},
  number={2},
  pages={1174--1185},
  doi={10.1109/TCSVT.2023.3292285}}

@article{tcsvt_selfsemi,
  title={Self Pseudo Entropy Knowledge Distillation for Semi-supervised Semantic Segmentation},
  author={Lu, Xiaoqiang and Jiao, Licheng and Li, Lingling and Liu, Fang and Liu, Xu and Yang, Shuyuan},
  journal={IEEE Transactions on Circuits and Systems for Video Technology},
  year={2024},
  volume={34},
  number={8},
  pages={7359--7372},
  publisher={IEEE}
}

@article{open_tcsvt,
  author={Pan, Yuwen and Sun, Rui and Wang, Yuan and Yang, Wenfei and Zhang, Tianzhu and Zhang, Yongdong},
  journal={IEEE Transactions on Circuits and Systems for Video Technology}, 
  title={Purify Then Guide: A Bi-Directional Bridge Network for Open-Vocabulary Semantic Segmentation}, 
  year={2025},
  volume={35},
  number={1},
  pages={343--356},
  doi={10.1109/TCSVT.2024.3464631}}

@article{bssnet,
  title={{BSSNet}: A Real-Time Semantic Segmentation Network for Road Scenes Inspired from AutoEncoder},
  author={Shi, Xiaoqiang and Yin, Zhenyu and Han, Guangjie and Liu, Wenzhuo and Qin, Li and Bi, Yuanguo and Li, Shurui},
  journal={IEEE Transactions on Circuits and Systems for Video Technology},
  year={2024},
  volume={34},
  number={5},
  pages={3424--3438},
  publisher={IEEE}
}

@article{adaptiveocc,
  author={Yang, Tianyu and Qian, Yeqiang and Yan, Weihao and Wang, Chunxiang and Yang, Ming},
  journal={IEEE Transactions on Circuits and Systems for Video Technology}, 
  title={{AdaptiveOcc}: Adaptive Octree-Based Network for Multi-Camera {3D} Semantic Occupancy Prediction in Autonomous Driving}, 
  year={2025},
  volume={35},
  number={3},
  pages={2173--2187},
  doi={10.1109/TCSVT.2024.3492289}
  }

@inproceedings{unimatch,
  title={Revisiting Weak-to-Strong Consistency in Semi-Supervised Semantic Segmentation},
  author={Yang, Lihe and Qi, Lei and Feng, Litong and Zhang, Wayne and Shi, Yinghuan},
  booktitle={Proceedings of the IEEE/CVF Conference on Computer Vision and Pattern Recognition},
  pages={7236--7246},
  year={2023}
}

@inproceedings{semivl,
  title={{SemiVL}: Semi-Supervised Semantic Segmentation with Vision-Language Guidance},
  author={Hoyer, Lukas and Tan, David Joseph and Naeem, Muhammad Ferjad and Van Gool, Luc and Tombari, Federico},
  booktitle={European Conference on Computer Vision},
  pages={257--275},
  year={2024}
}

@article{unimatchv2,
  title={{UniMatch V2}: Pushing the Limit of Semi-Supervised Semantic Segmentation},
  author={Yang, Lihe and Zhao, Zhen and Zhao, Hengshuang},
  journal={IEEE Transactions on Pattern Analysis and Machine Intelligence},
  year={2025},
  volume={47},
  number={4},
  pages={3031--3048},
  doi={10.1109/TPAMI.2025.3528453}
}

@inproceedings{allspark,
  title={{AllSpark}: Reborn Labeled Features from Unlabeled in Transformer for Semi-Supervised Semantic Segmentation},
  author={Wang, Haonan and Zhang, Qixiang and Li, Yi and Li, Xiaomeng},
  booktitle={Proceedings of the IEEE/CVF Conference on Computer Vision and Pattern Recognition},
  pages={3627--3636},
  year={2024}
}

@ARTICLE{perturbmatch,
  author={Zhang, Guoqing and Han, Chenxing and Chen, Yadang and Sun, Le and Cao, Yulin and Zheng, Yuhui},
  journal={IEEE Transactions on Circuits and Systems for Video Technology}, 
  title={{PerturbMatch}: Multi-level Progressive Perturbation for Semi-Supervised Semantic Segmentation}, 
  year={2026},
  note={early access, doi: 10.1109/TCSVT.2026.3717713},
  doi={10.1109/TCSVT.2026.3717713},
  }

@inproceedings{pidnet,
  title={{PIDNet}: A Real-Time Semantic Segmentation Network Inspired by {PID} Controllers},
  author={Xu, Jiacong and Xiong, Zixiang and Bhattacharyya, Shankar P.},
  booktitle={Proceedings of the IEEE/CVF Conference on Computer Vision and Pattern Recognition},
  pages={19529--19539},
  year={2023}
}

@inproceedings{ssmis,
  title         = {Beyond Random Sampling: Distribution-Aware Alignment for Semi-Supervised Medical Image Segmentation},
  author        = {Yan, Weihao and Qian, Yeqiang and Dong, Yi and Yang, Ming},
  booktitle     = {Proceedings of the European Conference on Computer Vision},
  year          = {2026},
  eprint        = {2607.04249},
  archivePrefix = {arXiv},
  primaryClass  = {cs.CV},
  url           = {https://arxiv.org/abs/2607.04249}
}

@inproceedings{k-center,
    title={Active Learning for Convolutional Neural Networks: A Core-Set Approach},
    author={Ozan Sener and Silvio Savarese},
    booktitle={International Conference on Learning Representations},
    year={2018},
}

@inproceedings{typiclust,
  title = {Active Learning on a Budget: Opposite Strategies Suit High and Low Budgets},
  author = {Hacohen, Guy and Dekel, Avihu and Weinshall, Daphna},
  booktitle = {Proceedings of the 39th International Conference on Machine Learning},
  pages = {8175--8195},
  year = {2022},
  volume = {162},
  publisher = {PMLR},
}

@inproceedings{badge,
  title={Deep Batch Active Learning by Diverse, Uncertain Gradient Lower Bounds},
  author={Ash, Jordan T. and Zhang, Chicheng and Krishnamurthy, Akshay and Langford, John and Agarwal, Alekh},
  booktitle={International Conference on Learning Representations},
  year={2020}
}

@InProceedings{entropy,
  title = 	 {Deep {B}ayesian Active Learning with Image Data},
  author =       {Yarin Gal and Riashat Islam and Zoubin Ghahramani},
  booktitle = 	 {Proceedings of the 34th International Conference on Machine Learning},
  pages = 	 {1183--1192},
  year = 	 {2017},
  volume = 	 {70},
  publisher =    {PMLR},
}

@article{dwba,
  title={Dynamic weighting and boundary-aware active domain adaptation for semantic segmentation in autonomous driving environment},
  author={Guan, Licong and Yuan, Xue},
  journal={IEEE Transactions on Intelligent Transportation Systems},
  volume={25},
  number={11},
  pages={18461--18471},
  year={2024},
  publisher={IEEE}
}

@article{bada,
  title={Boundary-based active domain adaptation for semantic segmentation under adverse conditions},
  author={Xu, Xianzhe and Yen, Gary G and Zhao, Chaoqiang and Sun, Qiyu and Ren, Wenqi and Sheng, Lu and Tang, Yang},
  journal={IEEE Transactions on Neural Networks and Learning Systems},
  volume={36},
  number={8},
  pages={14721--14734},
  year={2025},
  publisher={IEEE}
}

@inproceedings{gao_superpixel,
  title={Efficient active domain adaptation for semantic segmentation by selecting information-rich superpixels},
  author={Gao, Yuan and Wang, Zilei and Zhang, Yixin and Tu, Bohai},
  booktitle={European Conference on Computer Vision},
  pages={405--421},
  year={2024},
  organization={Springer}
}

@inproceedings{ripu,
  title={Towards Fewer Annotations: Active Learning via Region Impurity and Prediction Uncertainty for Domain Adaptive Semantic Segmentation},
  author={Xie, Binhui and Yuan, Longhui and Li, Shuang and Liu, Chi Harold and Cheng, Xinjing},
  booktitle={Proceedings of the IEEE/CVF Conference on Computer Vision and Pattern Recognition},
  pages={8068--8078},
  year={2022}
}

@inproceedings{d2ada,
  title={{D2ADA}: Dynamic Density-Aware Active Domain Adaptation for Semantic Segmentation},
  author={Wu, Tsung-Han and Liou, Yi-Syuan and Yuan, Shao-Ji and Lee, Hsin-Ying and Chen, Tung-I and Huang, Kuan-Chih and Hsu, Winston H.},
  booktitle={European Conference on Computer Vision},
  pages={449--467},
  year={2022}
}

@inproceedings{halo,
  title={Hyperbolic Active Learning for Semantic Segmentation under Domain Shift},
  author={Franco, Luca and Mandica, Paolo and Kallidromitis, Konstantinos and Guillory, Devin and Li, Yu-Teng and Darrell, Trevor and Galasso, Fabio},
  booktitle={Proceedings of the 41st International Conference on Machine Learning},
  pages={13864--13884},
  volume={235},
  publisher={PMLR},
  year={2024}
}

@ARTICLE{ning2023madav2,
  author={Ning, Munan and Lu, Donghuan and Xie, Yujia and Chen, Dongdong and Wei, Dong and Zheng, Yefeng and Tian, Yonghong and Yan, Shuicheng and Yuan, Li},
  journal={IEEE Transactions on Pattern Analysis and Machine Intelligence}, 
  title={{MADAv2}: Advanced Multi-Anchor Based Active Domain Adaptation Segmentation}, 
  year={2023},
  volume={45},
  number={11},
  pages={13553--13566},
  }

@article{tcsvt_active,
  title={Active learning based {3D} semantic labeling from images and videos},
  author={Rong, Mengqi and Cui, Hainan and Hu, Zhanyi and Jiang, Hanqing and Liu, Hongmin and Shen, Shuhan},
  journal={IEEE Transactions on Circuits and Systems for Video Technology},
  year={2022},
  volume={32},
  number={12},
  pages={8101--8115},
  publisher={IEEE}
}

@article{dcov,
  title={Dynamic Confidence Variance for Generalized Coreset in Active Learning},
  author={Wan, Tianjiao and Gao, Zijian and Gong, Xudong and Feng, Dawei and Zhang, Xingxing and Ding, Bo and Wang, Yijie and Wang, Huaimin and Xu, Kele},
  journal={IEEE Transactions on Circuits and Systems for Video Technology},
  year={2026},
  volume={36},
  number={3},
  pages={3231--3245},
  doi={10.1109/TCSVT.2025.3622313}
}

@article{adssemiseg,
  title={Adversarial Dual-Student With Differentiable Spatial Warping for Semi-Supervised Semantic Segmentation},
  author={Cao, Cong and Lin, Tianwei and He, Dongliang and Li, Fu and Yue, Huanjing and Yang, Jingyu and Ding, Errui},
  journal={IEEE Transactions on Circuits and Systems for Video Technology},
  year={2023},
  volume={33},
  number={2},
  pages={793--803},
  doi={10.1109/TCSVT.2022.3206496}
}

@article{dip,
  title={Domain-Invariant Prototypes for Semantic Segmentation},
  author={Yang, Zhengeng and Yu, Hongshan and Sun, Wei and Cheng, Li and Mian, Ajmal},
  journal={IEEE Transactions on Circuits and Systems for Video Technology},
  year={2024},
  volume={34},
  number={8},
  pages={7614--7627},
  doi={10.1109/TCSVT.2024.3375306}
}

@inproceedings{semidavil,
  title={{SemiDAViL}: Semi-supervised domain adaptation with vision-language guidance for semantic segmentation},
  author={Basak, Hritam and Yin, Zhaozheng},
  booktitle={Proceedings of the Computer Vision and Pattern Recognition Conference},
  pages={9816--9828},
  year={2025}
}
\vspace{-12.5 mm}
\begin{IEEEbiography}[{\includegraphics[width=1in,height=1.2in,clip,keepaspectratio]{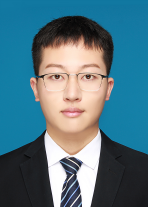}}]{\textbf{Weihao Yan}}
received a Bachelor's degree in Automation from Shanghai Jiao Tong University, Shanghai, China, in 2020. He is working towards a Ph.D. degree in Control Science and Engineering from Shanghai Jiao Tong University.\\
His main fields of interest are autonomous driving systems, computer vision, and domain adaptation. His current research activities include virtual-to-real transfer learning, scene segmentation, and foundation models.
\end{IEEEbiography}
\vspace{-12.5 mm}
\begin{IEEEbiography}[{\includegraphics[width=1in,height=1.2in,clip,keepaspectratio]{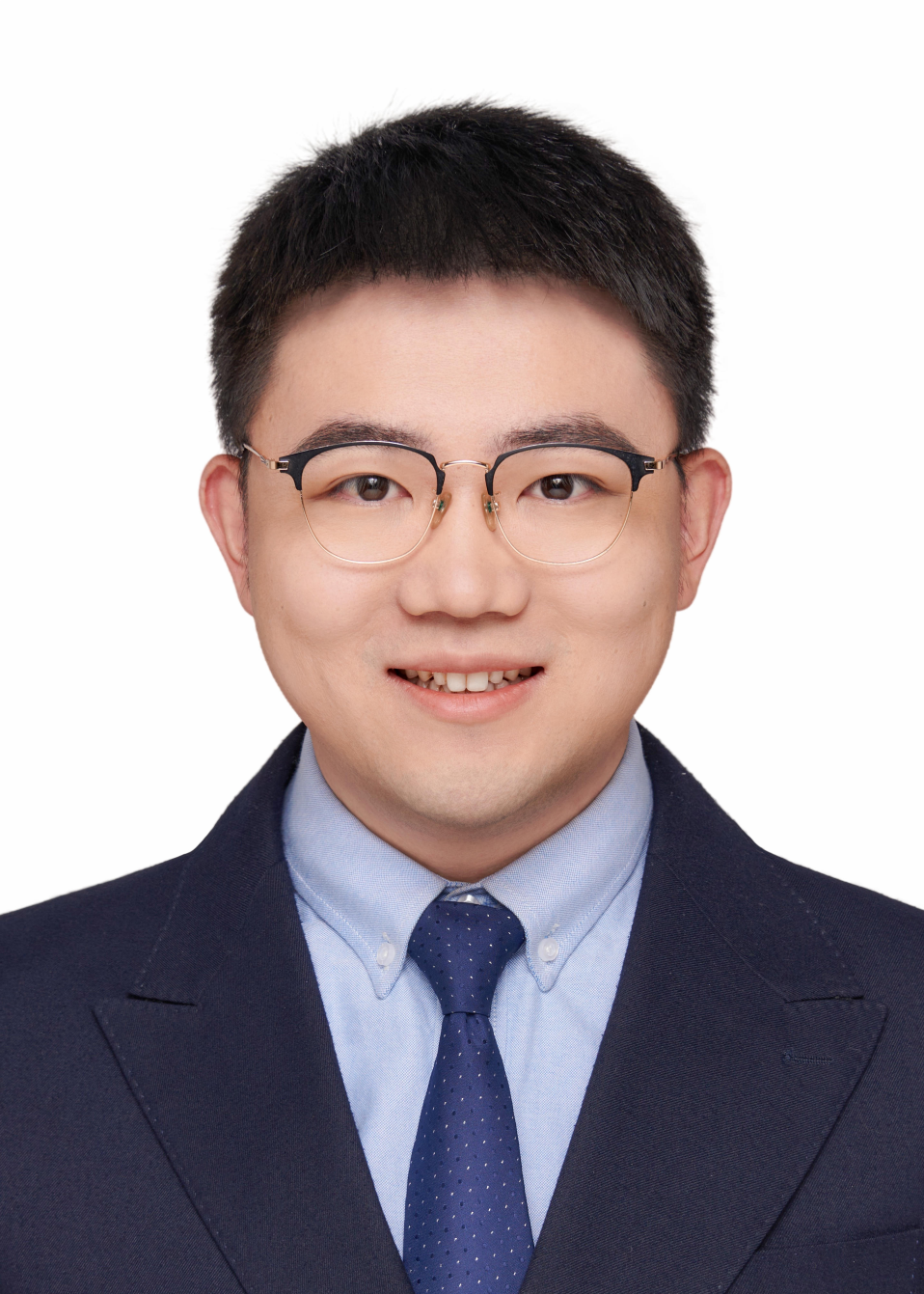}}]{\textbf{Yeqiang Qian}}
received a Ph.D. degree in control science and engineering from Shanghai Jiao Tong University, Shanghai, China, in 2020. He is currently a Tenure Track Associate Professor with the Department of Automation at Shanghai Jiao Tong University, Shanghai, China. \\ 
His main research interests include computer vision, pattern recognition, machine learning, and their applications in intelligent transportation systems.
\end{IEEEbiography}
\vspace{-12.5 mm}
\begin{IEEEbiography}[{\includegraphics[width=1in,height=1.25in,clip,keepaspectratio]{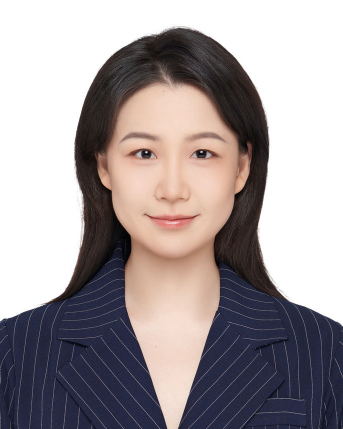}}]{Yueyuan Li} 
received a Bachelor's degree in Electrical and Computer Engineering from the University of Michigan-Shanghai Jiao Tong University Joint Institute, Shanghai, China, in 2020. She is working towards a Ph.D. degree in Control Science and Engineering from Shanghai Jiao Tong University.\\
Her main fields of interest are the security of the autonomous driving system and driving decision-making. Her current research activities include driving decision-making models, driving simulation, and virtual-to-real model transfer.
\end{IEEEbiography}
\vspace{-12.5 mm}
\begin{IEEEbiography}[{\includegraphics[width=1in,height=1.25in,clip,keepaspectratio]{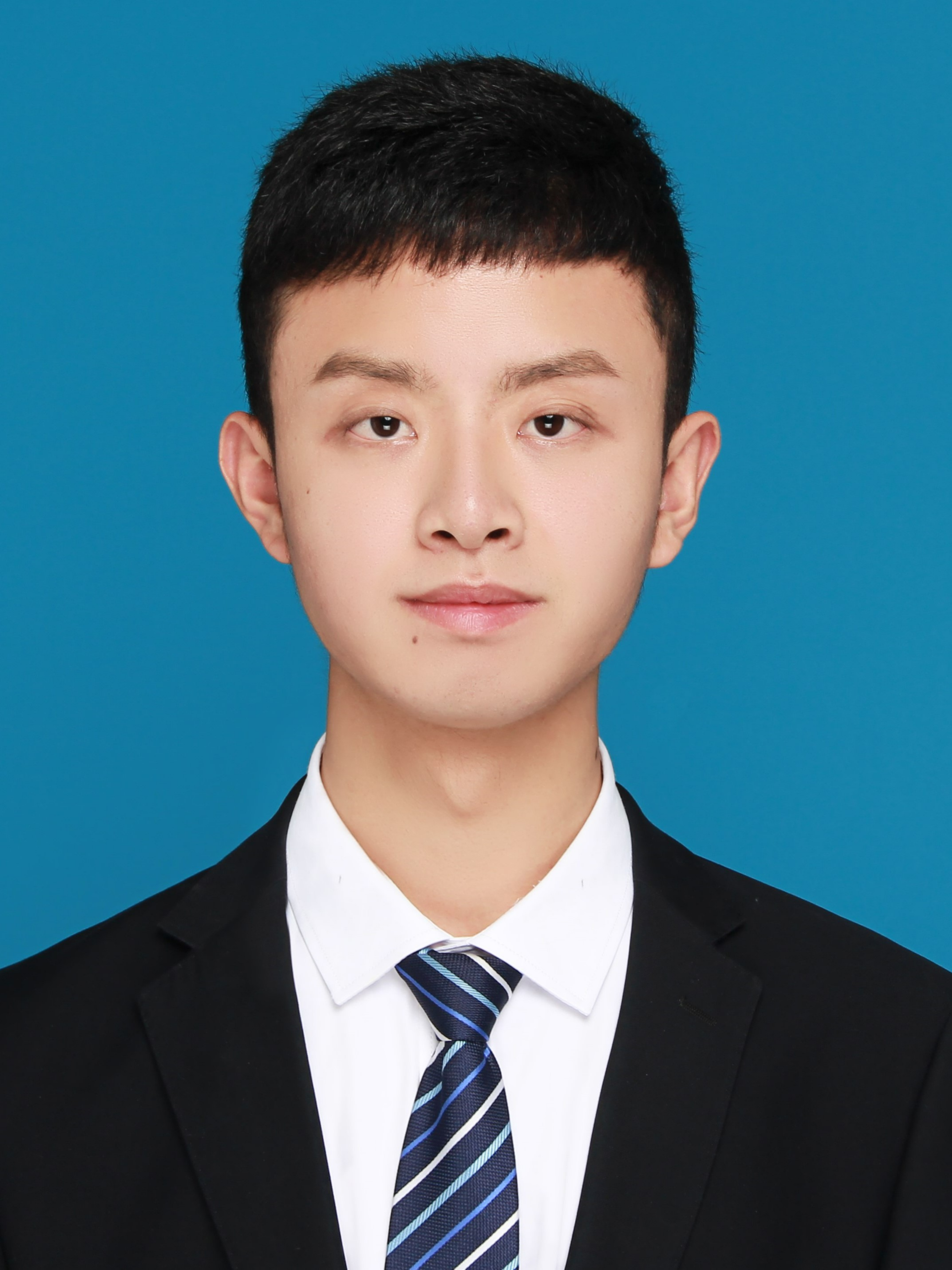}}]{Tao Li} 
received a Bachelor's degree in Automation from Shanghai Jiao Tong University, Shanghai, China, in 2020 and a Ph.D. degree in Control Science and Engineering from Shanghai Jiao Tong University in 2025. \\
His main research interests include machine learning and optimization methods.
\end{IEEEbiography}
\vspace{-12.5 mm}
\begin{IEEEbiography}[{\includegraphics[width=1in,height=1.2in,clip,keepaspectratio]{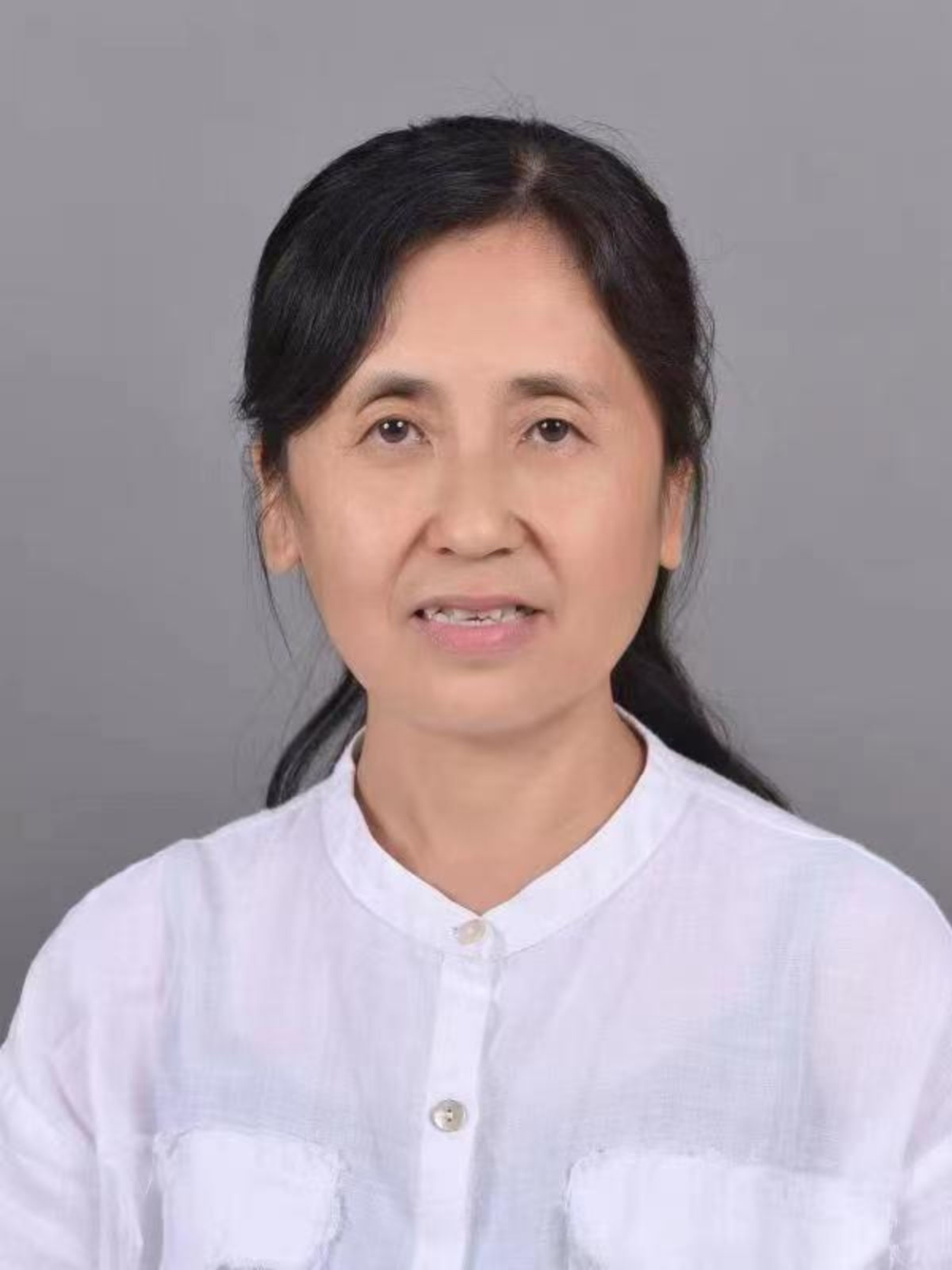}}]{\textbf{Chunxiang Wang}}
received the Ph.D. degree in mechanical engineering from the Harbin Institute of Technology, Harbin, China, in 1999.\\
She is currently an Associate Professor with the Department of Automation at Shanghai Jiao Tong University, Shanghai, China. Her research interests include robotic technology and electromechanical integration.
\end{IEEEbiography}
\vspace{-12.5 mm}
\begin{IEEEbiography}[{\includegraphics[width=1in,height=1.2in,clip,keepaspectratio]{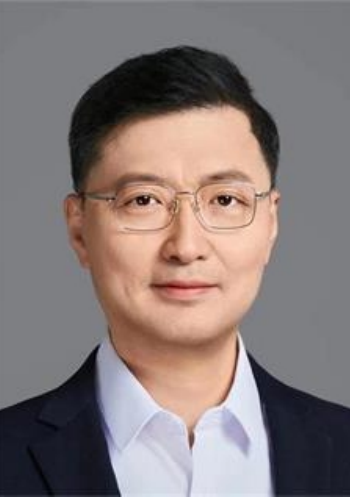}}]{\textbf{Ming Yang}}
received the Master and Ph.D. degrees from Tsinghua University, Beijing, China, in 1999 and 2003, respectively.\\
He is currently the Full Tenure Professor at Shanghai Jiao Tong University, the deputy director of the Innovation Center of Intelligent Connected Vehicles. He has been working in the field of intelligent vehicles for more than 20 years. He participated in several related research projects, such as the THMR-V project (first intelligent vehicle in China), European CyberCars and CyberMove projects, CyberC3 project, CyberCars-2 project, ITER transfer cask project, AGV, etc.
\end{IEEEbiography}


\end{document}